\documentclass[10pt,twocolumn,letterpaper]{article}

\usepackage[pagenumbers]{wacv} 

\definecolor{wacvblue}{rgb}{0.21,0.49,0.74}
\usepackage[pagebackref,breaklinks,colorlinks,allcolors=wacvblue]{hyperref}

\usepackage{graphicx}
\usepackage{pifont}
\usepackage{xcolor}

\newcommand{\cmark}{\textcolor{green!60!black}{\ding{51}}} 
\newcommand{\xmark}{\textcolor{red!70!black}{\ding{55}}} 

\usepackage{booktabs}
\usepackage{microtype}
\usepackage{tabularx} 
\usepackage{multirow} 
\usepackage{orcidlink}
\usepackage{array}
\usepackage{enumitem}
\usepackage{pgfplots}
\usepackage{dblfloatfix}
\pgfplotsset{compat=1.18}
\def\wacvPaperID{779} 
\def\confName{WACV}
\def\confYear{2027}

\title{JOPP-3D: Joint Open Vocabulary Semantic Segmentation \\ on Point Clouds and Panoramas}

\author{Sandeep Inuganti$^{1,2}$ \and
Hideaki Kanayama$^{3}$ \and
Kanta Shimizu$^{3,4}$ \and Mahdi Chamseddine$^{1,2}$ \and Soichiro Yokota$^{3,4}$ \and Didier Stricker$^{1,2}$ \and Jason Rambach$^{1}$ \and
$^{1}${\small German Research Center for Artificial Intelligence, DFKI, Germany} 
$^{2}${\small RPTU Kaiserslautern, Germany} $^{3}${\small Ricoh Company, Ltd. Japan}\\
$^{4}${\small Ricoh International B.V. - Niederlassung Deutschland, Germany}\\
$^{1}${\tt\small \{firstname.lastname\}@dfki.de} $^{3}${\tt\small \{firstname.lastname\}@jp.ricoh.com}}

\begin{document}
\maketitle
\begin{abstract}
Semantic segmentation across visual modalities such as 3D point clouds and panoramic images remains a challenging task, primarily due to the scarcity of annotated data and the limited adaptability of fixed-label models. In this paper, we present JOPP-3D, an open-vocabulary segmentation framework that jointly leverages panoramic and point cloud data to enable language-driven scene understanding.
We convert RGB-D panoramic images into their corresponding wide field-of-view tangential perspectives and 3D point clouds, then use these modalities to extract and align foundational vision-language features. This allows natural language querying to generate semantic masks on both input modalities. Experimental evaluation on the Stanford-2D-3D-s and ToF-360 datasets demonstrates the capability of JOPP-3D to produce coherent and semantically meaningful segmentations across panoramic and 3D domains. Our proposed method achieves a significant improvement compared to the SOTA in open and closed vocabulary 2D and 3D semantic segmentation. Code: \href{https://github.com/dfki-av/JOPP-3D}{github.com/dfki-av/JOPP-3D}
\end{abstract}

\section{Introduction}
Semantic understanding of complex real-world environments is a fundamental requirement for autonomous systems and robotics.
Traditional semantic segmentation approaches rely heavily on large-scale annotated datasets~\cite{cordts2016cityscapes}, which are often infeasible to obtain and label in unstructured and dynamically changing domains.
Moreover, existing segmentation methods are typically constrained to either 2D images or 3D point clouds and depend on predefined class sets that limit their generalization to new object categories~\cite{qi2017pointnet, thomas2019kpconv}. These limitations hinder the 
adaptability of perception systems in open and evolving environments where new objects frequently appear, and exhaustive annotation is impractical.
\begin{figure}
    \centering
    \captionsetup{type=figure}
        \includegraphics[width=0.9\linewidth]{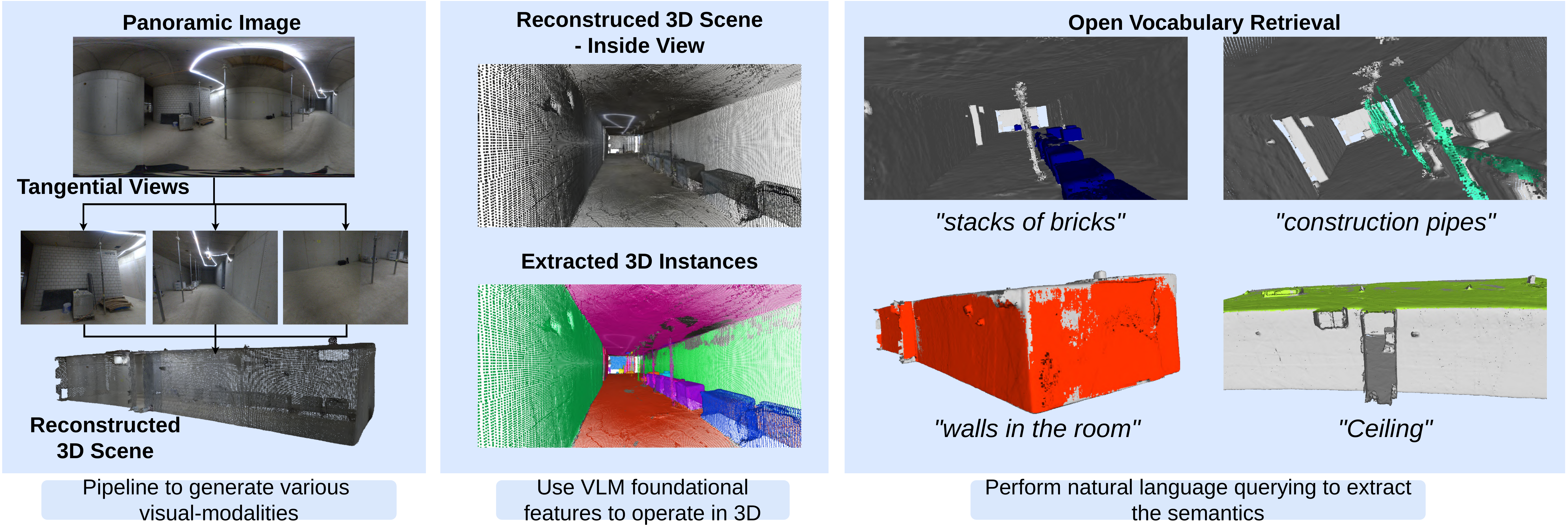}
        \captionof{figure}{Joint treatment of panoramas and 3D point clouds in our proposed pipeline. Example of zero-shot performance on a construction site scene with varying mask scale.}
    \label{fig:teaser}
    \vspace{-0.7cm}
\end{figure}

Recent advances in vision-language models (VLMs) have shown the potential of open-vocabulary segmentation, enabling models to recognize and localize arbitrary objects through natural language queries without explicit training labels. Foundational works such as CLIP~\cite{radford2021learning} have demonstrated strong cross-modal alignment between visual and textual features, inspiring open-vocabulary extensions to image segmentation tasks~\cite{luddecke2022image, li2022languagedriven}. However, most existing works focus exclusively on perspective images and do not generalize well to panoramic imagery or 3D point clouds, both of which are essential for comprehensive spatial understanding. Panoramic images offer complete 360° coverage~\cite{coors2018spherenet, zheng2024open}, while 3D point clouds provide geometric fidelity~\cite{kolodiazhnyi2024oneformer3d, peng2023openscene}. However, current research has not yet addressed the joint semantic interpretation of these modalities in an open-vocabulary context. Furthermore, aligning 2D panoramic domains and 3D spatial representations remains a major challenge due to geometric distortions, sparse depth estimation, and modality gaps. 

To address these challenges, we propose a unified framework for open-vocabulary segmentation of 3D point clouds and panoramic images, leveraging large pre-trained models~\cite{segmentAnything, radford2021learning}
to achieve language-driven segmentation. Our approach bridges 2D-3D understanding by constructing 3D point clouds of a scene directly from its multi-view panoramic captures - enabling consistent semantic prediction across both modalities. The main contributions of this work are summarized as follows:

\begin{itemize}

    \item An open-vocabulary approach that leverages both panoramic images and their 3D point cloud reconstructions to perform semantic segmentation.
    \item To maintain compatibility with vision-language models, an effective pipeline for adapting panoramic inputs by designing wide-field coverage tangential decomposition of panoramas.
    \item A 3D to panoramic semantic label propagation method via a depth correspondence strategy to enable multi-view-consistent semantic maps.
    \item Reliable open-vocabulary retrieval of large structures like walls, floors, ceilings, due to wide-FoV tangential views.
\end{itemize}


\noindent Overall, we tackle open-vocabulary based 3D segmentation by processing panoramic images and their 3D point clouds jointly -  making panoramic and 3D representations equivalent inputs to the same pipeline (see Figure~\ref{fig:teaser} and supplementary sec.C ). We propose two model paradigms that differ in the supervision required for instance proposals: \textbf{JOPP-3D}, one that only requires supervision for instance masks, and \textbf{JOPP-3D(f)}, that is foundation model-based - demonstrating the feasibility of open-vocabulary segmentation. 
\section{Related Work}
\subsection{3D Segmentation}
3D scene segmentation aims to assign semantic~\cite{armeni20163d} or instance-level~\cite{rozenberszki2022language} labels to each point in a 3D point cloud, thereby enabling detailed understanding of object geometry and spatial layout in complex environments. 
The introduction of deep learning revolutionized point cloud processing. PointNet~\cite{qi2017pointnet} and PointNet++~\cite{qi2017pointnet++} pioneered direct learning on point sets by aggregating point-wise and local neighborhood features, eliminating the need for voxelization. 
Transformer-based architectures, including PointTransformer and its variants~\cite{zhao2021point, wu2022point, wu2024point, chang2024mikasa}, further enhanced contextual reasoning. Other research trends have explored multimodal and cross-domain 3D segmentation, integrating complementary data such as multi-view images~\cite{jaritz2019multi, wang2026inpaint360gs} or leveraging self-supervised and contrastive learning for pretraining~\cite{hou2021exploring, knaebel2025ditr}. 

Despite these advances, most existing frameworks~\cite{choy20194d, wang2023octformer} are constrained by fixed taxonomies and the necessity of large-scale annotated datasets, which are difficult to obtain in unstructured real-world environments. In contrast, open-vocabulary 3D segmentation seeks to overcome these limitations by coupling 3D perception models with vision-language representations~\cite{peng2023openscene, Jiang_2024_CVPR, nguyen2024open3dis, ovsam3d}. By leveraging pre-trained models such as CLIP~\cite{radford2021learning}, these approaches enable segmentation guided by natural language queries, allowing for zero-shot generalization to unseen categories.

\subsection{Handling Panoramic Imagery}
Unlike conventional perspective images, panoramic or equirectangular images capture the full field of view, making them particularly useful for robotics, autonomous driving, and construction monitoring~\cite{yang2020ds, wen2024panacea, kanayama2025tof}. However, their inherent geometric distortions, wide spatial coverage, and large variations in object scale pose significant challenges for traditional segmentation networks designed for planar projections. To address these issues, specialized architectures~\cite{teng2024360bev, guttikonda2024SFSS, zhang2024behind} were proposed, enabling more consistent \& geometry-aware feature extraction. Despite these advances, panoramic segmentation methods remain largely limited to supervised training and closed-set class definitions. 

On the other hand, foundation models such as SAM~\cite{segmentAnything} and CLIP~\cite{radford2021learning} have strong potential for understanding semantics in perspective images. However, directly applying them to panoramic images is non-trivial~\cite{chamseddine2026panosamic} due to geometric distortions. To tackle this,
there are three feasible approaches: polyhedral projection~\cite{Snyder1992AnEM}, Cubemap representation~\cite{Woo_Han_2020, cheng2018cube}, and deformable adapter network (DAN)~\cite{zheng2024open}. 
Cubemap projection unfolds a 360° panorama onto the six faces of a cube, providing individual views with a 90° field-of-view (FoV), however this results in artifacts from boundary discontinuities. Meanwhile, DAN from the RGB-only method OOOPS~\cite{zheng2024open} requires supervised training to provide perspective-based representations for their downstream open-vocabulary segmentation. 

In contrast, our method proposes a geometrically grounded panoramic decomposition, eliminating the need for supervised training to learn deformations. Our approach uses a polyhedral projection~\cite{Snyder1992AnEM} that approximates the sphere using a regular polyhedron and project the panorama onto its $F$ faces. This introduces inter-view overlap, mitigating boundary discontinuities. Approaches like Eder \etal~\cite{Eder_2020_CVPR} pioneered ~\cite{Snyder1992AnEM} via tangent images, but their FoV operates around $73^\circ$ per image, constraining contextual coverage. To improve the tangential decomposition, we propose a perspective-projection-based approach, which allows for a wider FoV of $100^\circ$. Furthermore, by incorporating depth information and operating in 3D, our approach produces semantically consistent panoramic segmentation.

\subsection{Open-vocabulary scene understanding}
VLMs have recently emerged as a powerful paradigm for enabling open-vocabulary and language-guided perception in computer vision. 
The seminal model, CLIP~\cite{radford2021learning}, and the subsequent efforts~\cite{jia2021scaling, li2022blip}, establish a semantic bridge between images and natural language that generalizes effectively to downstream tasks.
\begin{figure*}
    \centering
    \captionsetup{type=figure}
        \includegraphics[width=0.8\linewidth]{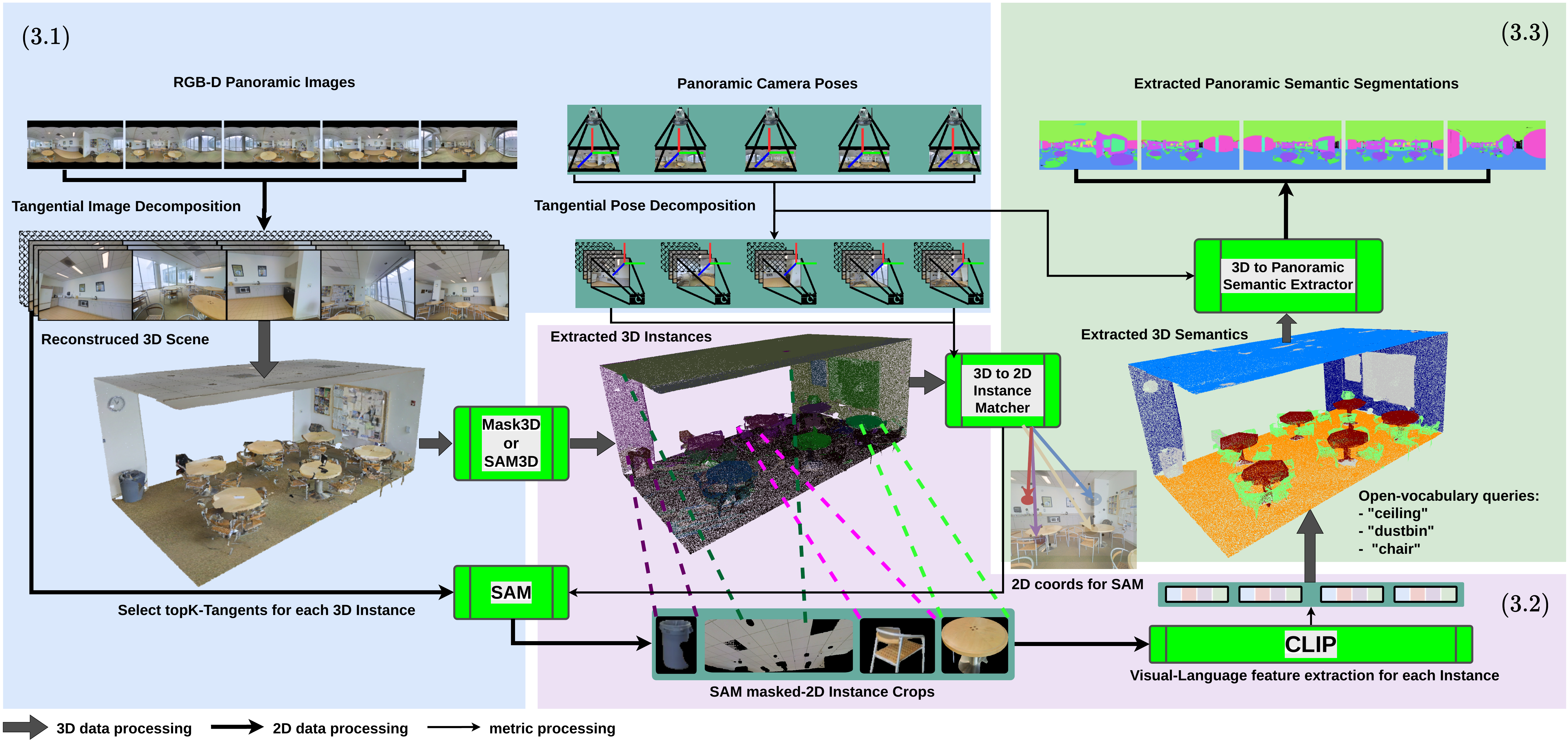}
        \captionof{figure}{Methodology: 
        Sec.$(3.1)$ We decompose all the panoramic captures of a scene into their corresponding tangential perspectives and depths. The other outputs of this step are also the tangential poses and the 3D point cloud. Sec.$(3.2)$ We then use these different visual modalities to extract 3D instances~\cite{schult2023mask3d, sam3d} of the reconstructed scene and align these instances with CLIP~\cite{radford2021learning} embeddings. Sec.$(3.3)$ In the final step we use open-vocabulary querying to extract the required 3D scene semantics, and further use a depth correspondence-based 3D-to-panoramic semantic extraction method, to obtain the semantic segmentation for each panoramic input.}
    \label{fig:method}
    \vspace{-0.5cm}
\end{figure*}
In 2D scene understanding, VLMs have been integrated into open-vocabulary object detection and segmentation~\cite{luddecke2022image, zhang2023simple, xu2023side, carion2025sam3segmentconcepts}.
Extending these ideas to 3D data has attracted significant attention. Recent works~\cite{peng2023openscene, fu2024scene} have explored mapping 3D representations into the same embedding space as pre-trained VLMs to perform open-vocabulary 3D scene understanding. These methods typically project 3D point features into the feature space learned by VLMs, enabling natural language-based querying and reasoning. Instead, to mitigate the noisy representations from per-point embeddings, we adapt aggregation of vision–language features per-instance, as proposed by OpenMask3D~\cite{takmaz2023openmask3d} - which result in more reliable embeddings. 
However, OpenMask3D~\cite{takmaz2023openmask3d} focuses on 3D instance segmentation using perspective RGB-D image sequences, while our work targets joint semantic segmentation of panoramic images and their reconstructed 3D point clouds. In contrast to instance-centric formulations, we address full scene-level segmentation, enabling recognition of both object instances (see supplementary sec.B) and large structural elements (e.g., \textit{walls} and \textit{floors}).
\newline

\noindent In summary, our work builds upon this growing body of research by extending VLM capabilities to both 3D point clouds and panoramic images, aiming to achieve unified scene understanding across visual modalities.

\section{Methodology}

Formally, for each scene we are provided with a set of panoramic images \{$\mathbf{I}^p \in \mathbb{R}^{W \times H \times 3}$\}, their depth maps \{$\mathbf{D}^p \in \mathbb{R}^{W \times H}$\}, and camera poses
\{$[\mathbf{R}^p, \mathbf{t}^p] \in SE(3)$\}. Each panoramic capture of this current scene is then decomposed into a set of $F$ perspectives:
\begin{equation}
\mathcal{F}_{\text{TD}}(\mathbf{I}^p, \mathbf{D}^p, [\mathbf{R}^p, \mathbf{t}^p]) = \{(\mathbf{I}_r, \mathbf{D}_r, [\mathbf{R}_r, \mathbf{t}_r]) \}_{r=1}^{F}
\end{equation}
where \(\mathcal{F}_{\text{TD}}\) is our Tangential Decomposition process from which we reconstruct a unified 3D point cloud representation of the scene, $\mathcal{P}^{\text{3D}} = \{ \mathbf{p}_i \in \mathbb{R}^3 \}_{i=1}^{N}$ (output of Sec~\ref{sec:tgtdec}). Using $\mathcal{P}^{\text{3D}}$ and its corresponding set of tangential decompositions $\{\mathcal{F}_{\text{TD}}(\mathbf{I}^p, \mathbf{D}^p, [\mathbf{R}^p, \mathbf{t}^p])\}$, we extract semantic labels $q_i$ and construct semantic point cloud $\mathcal{S}^{\text{3D}} = \{ (\mathbf{p}_i, q_i)\}_{i=1}^{N}$ in Sec~\ref{sec:semalign}. Finally in Sec~\ref{sec:3dtoP}, the output is a set of dense semantic maps $\{\mathbf{S}^p \in \mathbb{R}^{W \times H}\}$ corresponding to $\mathcal{S}^{\text{3D}}$.

To this end, as shown in~\cref{fig:method}, we design a unified framework consisting of three major components, each addressing a key challenge in our problem setting.

\subsection{Tangential Decomposition}
\label{sec:tgtdec}

\begin{figure}
    \centering
    \captionsetup[subfigure]{labelformat=empty}

    \begin{subfigure}[b]{0.155\textwidth}  
        \centering
        \includegraphics[width=\textwidth]{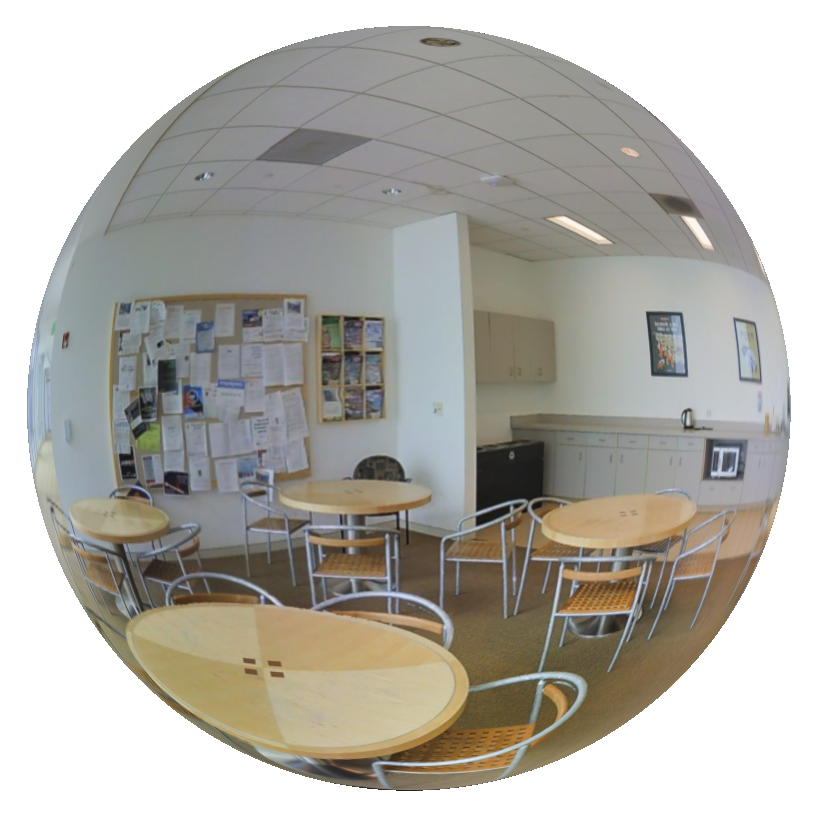}
    \end{subfigure}
    \begin{subfigure}[b]{0.155\textwidth}
        \centering
        \includegraphics[width=\textwidth]{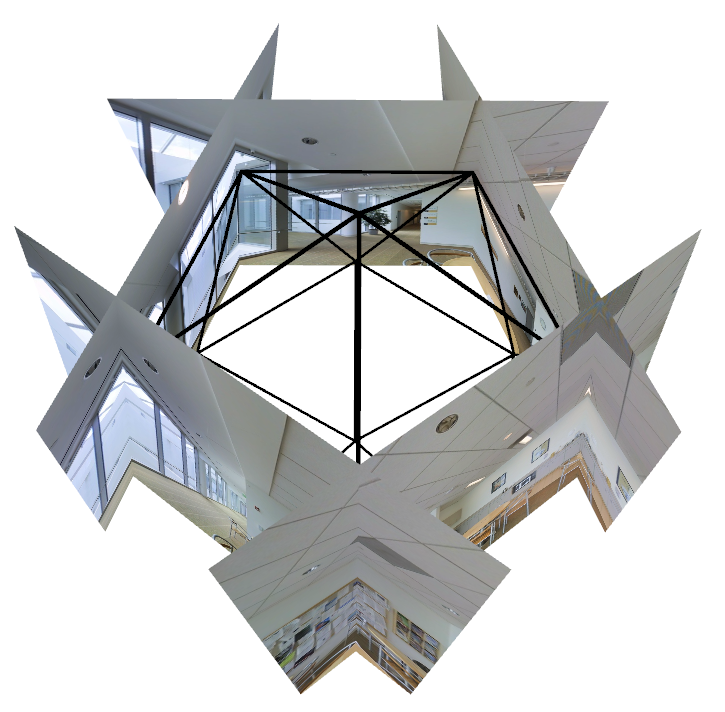}
    \end{subfigure}
    \begin{subfigure}[b]{0.155\textwidth}
        \centering
        \includegraphics[width=\textwidth]{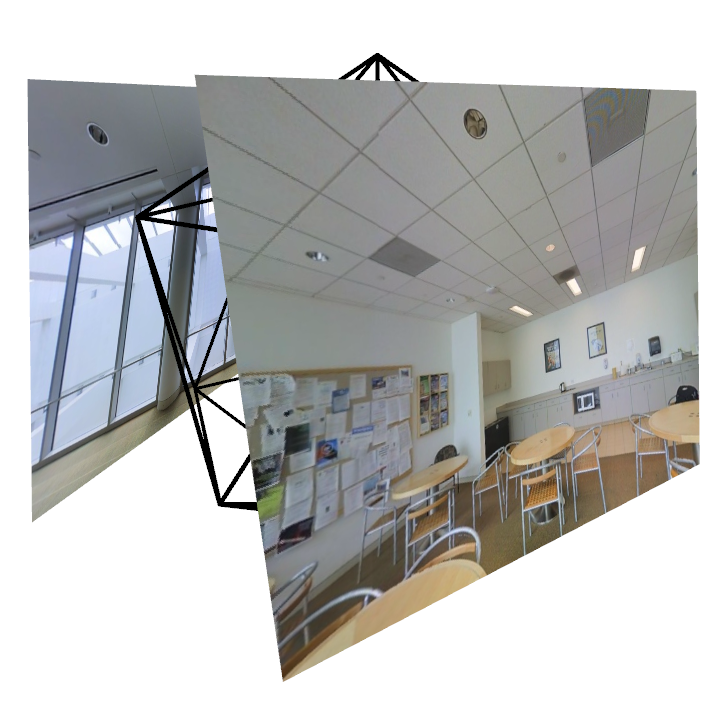}
    \end{subfigure}
    \caption{Representation of a spherical image using tangential decomposition. Left-right-bottom: Original spherical image;
    Tangential perspectives mapped onto the faces of an icosahedron; Some tangential perspective images extracted from the spherical image.}
    \label{fig:tangential_projection}
    \vspace{-0.5cm}
\end{figure}

As shown in Fig.~\ref{fig:tangential_projection}, we project the panoramic image \( \mathbf{I}^p \) and depth \( \mathbf{D}^p \) onto the 20 faces of a regular icosahedron, generating tangential decompositions \(\{\mathbf{I}_r\}_{r=1}^{20}\) and \(\{\mathbf{D}_r\}_{r=1}^{20}\) corresponding to distinct viewing directions. For each pixel $(i_r,j_r)$ in \(\mathbf{I}_r\) and \(\mathbf{D}_r\) of size $W \times H$, we compute its normalized camera‑space direction by applying the face‑specific rotation $\mathbf{R}_{r,\,\text{local}}$ to the local ray direction:
\begin{equation}
\label{eq:ray}
\begin{bmatrix}
x_r \\[4pt]
y_r \\[4pt]
z_r
\end{bmatrix}
=
\mathbf{R}_{r,\,\text{local}}
\begin{bmatrix}
(i_r - W/2)/f \\[4pt]
(j_r - H/2)/f \\[4pt]
1
\end{bmatrix}
\end{equation}

Now, the field-of-view, which we set to $100^\circ$ in our system to ensure a sufficiently wide coverage \& aspect ratio per tangent view. Our method exceeds the field-of-view achieved by Eder \etal~\cite{Eder_2020_CVPR}, while keeping the perspective projection within a geometrically stable range. Particularly, our field‑of‑view was designed such that:

(\textbf{i}) Every viewing direction is covered by at least $2$ tangential views, but with minimal distortion. (\textbf{ii}) Maintain the standard $4{:}3$ aspect ratio. This two-view redundancy reduces boundary fragmentation. Given (\textbf{i})\&(\textbf{ii}), our vertical and horizontal FoVs ($FOV_v, FOV_h$) are as follows: For a regular icosahedron, adjacent face normals are separated by \(\cos^{-1}(\sqrt{5}/3) \approx 41.8^\circ\).
Thus, from (\textbf{i}), $FOV_v \geq (2\times41.8)^\circ\approx 83.6^\circ$, and from (\textbf{ii}), $FOV_h=2\tan^{-1}\left(\frac{4}{3}\tan\left(\frac{FOV_v}{2}\right)\right)\approx 100.0^\circ.$ 

Next, the focal length $f$ is determined from the horizontal field of view $FOV_h$: $f = \frac{W/2}{\tan(FOV_h/2)}$. This ray in Eq.~\ref{eq:ray} is converted to spherical angles: $\theta_r = \arctan2(x_r,z_r)$, $\phi_r = \arctan2(y_r,\sqrt{x_r^2+z_r^2})$, and then mapped to equirectangular coordinates $(u_r,v_r)$:
\begin{equation}
    u_r = \frac{\theta_r+\pi}{2\pi}W_e,\quad v_r = \frac{\phi_r+\pi/2}{\pi}H_e
\end{equation}

Depth from the equirectangular map $\mathbf{D}^p(u_r,v_r)$ is corrected to Z-depth:
\begin{equation}
    \mathbf{D}^p_Z(u_r,v_r) = \mathbf{D}^p(u_r,v_r)\cdot \frac{z_r}{\sqrt{x_r^2+y_r^2+z_r^2}}
\end{equation}

Each pixel in the tangential perspective image is mapped to its corresponding location in the panorama:
$\mathbf{I}_r(i_r, j_r) = \mathbf{I}^p(u_r, v_r)$, $\mathbf{D}_r(i_r, j_r) = \mathbf{D}_Z^p(u_r, v_r)$.
RGB values are sampled using bilinear interpolation for smooth transitions, while depth values use nearest-neighbor interpolation for geometric accuracy. For each tangential face $r$, we reconstruct its local 3D point cloud and 
transform all points into the world coordinate system using the camera poses 
$[\mathbf{R}, \mathbf{t}]^p$:
\begin{equation}
\mathcal{P}^{\text{3D},\,\text{world}}_r
=
\mathbf{R}^p
\mathbf{D}^p_Z(u_r, v_r)
\begin{bmatrix}
x_{r}\\
y_{r}\\
z_{r}
\end{bmatrix}
+
\mathbf{t}^p 
\end{equation}

The full 3D point cloud of a panorama is obtained by aggregating all 20 faces. Finally, aggregating all panoramas in the scene and voxelizing yields the global colored 3D reconstruction:

\begin{equation}
\mathcal{P}^{\text{3D}}
=
\mathrm{Voxelize}
\left(
\bigcup_{p}
\;
\bigcup_{r=1}^{20}
\mathcal{P}^{\text{3D},\,\text{world}}_r
\right)
\end{equation}

\subsection{3D Instance Extraction and Semantic Alignment}
\label{sec:semalign}
Rather than relying on semantic class-specific supervision, our approach extracts instance-level 3D masks from an instance proposal backbone - enabling the segmentation pipeline to extract semantics from natural language queries. We consider two proposal sources that differ in the supervision they require: (1) \textbf{Mask3D}, an instance predictor trained with 3D instance masks labels; (2) \textbf{SAM3D}, which requires no 3D supervision at all, generating 3D instances from the 2D instance proposals of SAM~\cite{segmentAnything}. We refer the reader to Mask3D~\cite{schult2023mask3d} and SAM3D~\cite{sam3d} for further details on their respective frameworks. 

Given the point cloud from our reconstructed scene, \( \mathcal{P}^{\text{3D}} = \{ (x_i, y_i, z_i)\}_{i=1}^{N} \), 3D instance segmentation provides us with \(L\) instance proposals and per-point instance labels \({l_i \in \{1,...,L}\}\). Next, the point sets \(\mathcal{M}_j\) for each instance mask are obtained by grouping per instance label:
\begin{equation}
\mathcal{M}_j = \{ (x_i, y_i, z_i) \in \mathcal{P}^{\text{3D}} \mid l_i = j \}
\end{equation}





\subsubsection{Preparing 3D Instances for Semantic Alignment.}
The next step is to associate each instance mask \(\mathcal{M}_j\) with a corresponding \(d\)-dimensional semantic embedding \(\mathbf{e}_j^{\text{3D}} \in \mathbb{R}^d\) that enables open-vocabulary reasoning. The instance mask plays a crucial role in our framework: due to the wide field of view introduced by our tangential decomposition, each view may contain multiple objects and background regions. By isolating only the pixels belonging to the target instance, the mask prevents semantic contamination from surrounding objects and enables complete and consistent feature aggregation for the object of interest. To achieve this, we start by projecting the 3D points in \(\mathcal{M}_j\) into each tangential perspective image \(\mathbf{I}^t\) utilizing the tangential pose is given by $[\mathbf{R}^t, \mathbf{t}^t]=[\mathbf{R}_{r,\,\text{local}}\mathbf{R}^p, \mathbf{t}^p]$. 
Thus, we obtain the 3D to 2D matched pixel coordinates of instance \(\mathcal{M}_j\) in \(\mathbf{I}^t\) as:
\begin{equation}
\mathbf{M}^t_j(u,v) = \mathbf{K^t} (\, \mathbf{R}^t\, \mathbf{p} + \mathbf{t}^t),\quad\mathbf{p}\in\mathcal{M}_j
\end{equation}

where 
\(\mathbf{K}\) is the intrinsics of the tangential camera. 
Now we remove all the matches that are empty and only consider the set of non-empty matches: \(\{\mathbf{M}^t_j\}\). We then choose \(K\)-tangential images with the highest number of 3D to 2D matched pixel-coordinates by computing \(\textbf{top}_k(K)[\sum_{u_k,v_k}\mathbf{M}^t_j(u,v)]\) for each 3D instance mask \(\mathcal{M}_j\).



\subsubsection{2D Instance Segmentation using SAM and Semantic Alignment using CLIP.}
We have obtained a set of \(K\) tangential images \(\{\mathbf{I}^t\}_{k=1}^K\) for each \(\mathcal{M}_j\) and their corresponding \(K\) matched pixel-coordinates \(\{\mathbf{M}^t_j(u_k, v_k)\}\). We can now use them as reference points for SAM~\cite{segmentAnything} to compute the instance masks and their crops (bounding boxes), \(\mathbf{S}_{j,k}\), \(\mathbf{C}_{j,k}\) in 2D:
\begin{equation}
\mathbf{S}_{j,k}, \mathbf{C}_{j,k} = \text{SAM}(\mathbf{I}^t_k, \mathbf{M}^t_j(u_k, v_k))
\end{equation}
producing a consistent set of image crops highlighting the same 3D instance across $K$ tangential views. The instance crops \(\mathbf{C}_{j,k}\) are first masked with the instance masks \(\mathbf{S}_{j,k}\) and then passed through the CLIP image encoder~\cite{radford2021learning}. This masking of the crops is an important step while encoding instances - we elaborate more on this in Section~\ref{sec:ablations}. For each instance \( \mathcal{M}_j \), we compute its semantic embedding as the normalized average of its top-$K$ CLIP feature vectors:
\begin{equation}
\mathbf{e}_j^{\text{3D}} = \frac{1}{K} \sum_{k=1}^{K} \frac{\text{CLIP}(\mathbf{S}_{j,k} \odot \mathbf{C}_{j,k})}{\|\text{CLIP}(\mathbf{S}_{j,k} \odot \mathbf{C}_{j,k}) \|_2}
\end{equation}
This aggregated embedding \( \mathbf{e}_j^{\text{3D}} \in \mathbb{R}^d\) serves as the semantic descriptor of the 3D instance. With these embeddings, we repeatedly query using natural language for various classes, achieving 3D semantic segmentation (\(\mathcal{S}^{\text{3D}}\)) of the scene.



\subsection{3D to Panoramic Semantic Extraction}
\label{sec:3dtoP}
Once the 3D semantic segmentation has been obtained through language-guided alignment, it is possible to project these semantics back into the panoramic image domain - allowing the model to provide full-resolution semantic segmentation of panoramas, which are more interpretable and useful for downstream visual applications.

\subsubsection{Back-Projection from 3D to Panoramic Domain.}
Let the 3D semantic point cloud of the current scene, as previously defined be
\begin{equation}
\mathcal{S}^{\text{3D}} = \{ (\textbf{p}_i, q_i) \mid (x_i, y_i, z_i) \in \mathcal{P}^{\text{3D}} \}_{i=1}^{N}
\end{equation}
where \(\textbf{p}_i= (x_i, y_i, z_i) \in \mathbb{R}^3 \) denotes the 3D coordinates and \( q_i \in \{1,...,Q\} \) is the semantic label obtained from the $q^{\text{th}}$ open-vocabulary query, in a total of $Q$ different queries, else \( q_i = 0 \). For each panoramic image \( \mathbf{I}^p \), we have its corresponding camera pose represented by rotation \( \mathbf{R}^p \in SO(3) \) and translation \( \mathbf{t}^p \in \mathbb{R}^3 \). The panoramic depth map \( \mathbf{D}^p(u,v) \) provides the depth value of each pixel coordinate \( (u,v) \) in the panoramic projection space. We can transform each depth pixel into 3D coordinates in the global point cloud reference frame using:
\begin{equation}
\mathbf{X}^p(u,v) = \mathbf{R}^p \, \Pi^{-1}(u,v, \mathbf{D}^p(u,v)) + \mathbf{t}^p
\end{equation}
where \( \Pi^{-1}(\cdot) \) denotes the inverse panoramic projection function that maps spherical image coordinates to 3D space. This transformation aligns each panoramic depth point with the coordinate system of the global 3D semantic cloud.
\begin{figure}
    \centering
    \captionsetup{type=figure}
        \includegraphics[width=1.0\linewidth]{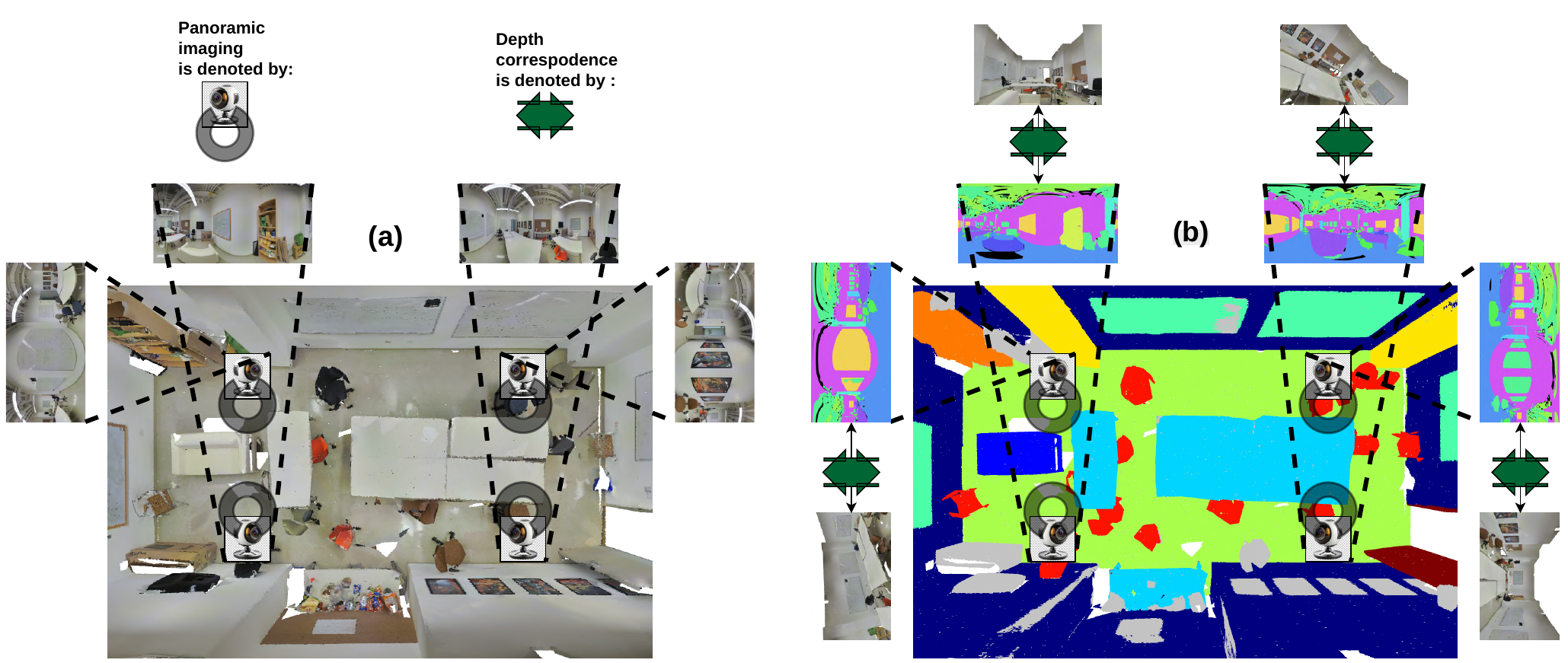}
        \captionof{figure}{3D-to-Panoramic Semantic Extractor. \textbf{{\fontfamily{phv}\selectfont (a)}} Shows the process of panoramic imaging from the top view of a Room. \textbf{{\fontfamily{phv}\selectfont (b)}} We try to emulate the same process on the extracted 3D semantic point cloud by placing the panoramic camera at the same poses. However, after each capture, we transfer the semantic labels of nearby Rooms (scenes) by establishing depth correspondences.}
    \label{fig:Pano}
    \vspace{-0.5cm}
\end{figure}
\subsubsection{Association via Nearest-Neighbor Matching.}
Once the transformed panoramic 3D points \( \mathbf{X}^p(u,v) \) are aligned with the 3D semantic point cloud \( \mathcal{S}^{\text{3D}} \), we associate to each pixel a semantic label based on its nearest 3D semantic neighbor \(\mathcal{N}\):
\begin{equation}
\mathcal{N}(\mathbf{X}^p(u,v)) = \arg \min_{(x_i, y_i, z_i) \in \mathcal{P}^{\text{3D}}} \| \mathbf{X}^p(u,v) - (x_i, y_i, z_i) \|_2
\end{equation}
The corresponding panoramic semantic label is then assigned as
\begin{equation}
q^p(u,v) = q_{i^*}\quad \text{where} \quad\{i^{*}|\mathcal{N}(\mathbf{X}^p(u,v)) = q_{i^*}\}
\end{equation}
This process produces a semantic map \(\mathbf{S}^p=\{ q^p(u,v) \}\) in the panoramic domain.

\subsubsection{Depth-Correspondence Consistency Across Scenes}
Direct nearest-neighbor assignment can lead to incomplete semantic projections, particularly near regions such as \textit{doorways} and \textit{corridors}, where panoramic views exhibit large depth discontinuities. To mitigate this, we introduce a depth correspondence strategy that leverages overlapping depth regions between neighboring panoramic scenes, as shown in Fig.~\ref{fig:Pano}. Given two adjacent (nearby) panoramas \( \mathbf{I}^p_1 \) and \( \mathbf{I}^p_2 \) with overlapping depth ranges, define a local correspondence set:
\begin{equation}
\Omega^p_{1,2} = \{ (u,v) \mid | \mathbf{D}^p_1(u,v) - \mathbf{D}^p_2(u',v') | < \delta \}
\end{equation}
where \( (u',v') \) denotes the pixel in \( \mathbf{I}^p_2 \) that projects to approximately the same 3D location as \( (u,v) \) in \( \mathbf{I}^p_1 \), and \( \delta \) is a small depth threshold that controls the tolerance of the correspondence. For each missing pixel value in \( \mathbf{S}_1^p \), we propagate the semantic embedding from \( \mathbf{S}_2^p \) if a valid correspondence exists:
\begin{equation}
q^p_1(u,v) = 
\begin{cases}
q^p_2(u',v'), & \text{if } (u,v) \in \Omega^p_{1,2} \text{ and } q^p_1(u,v) = 0 \\
q^p_1(u,v), & \text{otherwise}
\end{cases}
\end{equation}
This correspondence-guided refinement ensures semantic continuity across overlapping regions and maintains scene-level consistency in panoramic segmentation.
Further, an efficient implementation of the thresholding (\( \delta \)) for this refinement is elaborated in supplementary sec.D.

\section{Evaluation \& Analysis}



\subsection{Implementation Details}

For panoramic image processing, we generate \textbf{20} tangential perspective images of size $640 \times 480$ from each panoramic image of size $4096 \times 2048$. $K=5$ top tangents per instance; voxel size $=0.02$\,m. JOPP-3D and our OpenMask3D~\cite{takmaz2023openmask3d} runs, receive 3D instance proposals from a frozen Mask3D~\cite{schult2023mask3d} checkpoint pre-trained on the S3DIS dataset~\cite{armeni20163d}. JOPP-3D(f) uses our adaptation of SAM3D~\cite{sam3d} 
(more details provided in the supplementary sec.A). During evaluation, we use a single, fixed, language query for each dataset category, with no scene-specific prompting. In most cases, the query substitute is the same as dataset's category name. However, for certain ambiguous categories such as \textit{“board”}, we adopt a contextually specific substitute, \textit{“whiteboard”}, providing a clearer semantic ground. 
\begin{figure}[bp]
    \centering
    \begin{minipage}{0.41\linewidth}
        \centering
        \renewcommand{\arraystretch}{2.0}
        \resizebox{\linewidth}{!}{
        \begin{tabular}{l|cccc}
            Face Div. [$F$] & 6 & 8 & 12 & 20 \\
            \midrule
            $FOV_h$ [$^\circ$] & 180.0 & 150.3 & 138.9 & 100.0 \\
            \midrule
            $D(FOV_h)$ [$\downarrow$] & - & 15.2 & 8.1 & 2.4 \\
        \end{tabular}}
        \captionof{table}{$FOV_h$ vs. \\ boundary distortion factor.}
        \label{tab:dfov}
    \end{minipage}
    \hfill
    \begin{minipage}{0.58\linewidth}
        \centering
        \tiny
        \begin{tikzpicture}
        \begin{axis}[
            width=5.5cm,
            height=3cm,
            axis x line*=top,
            xlabel near ticks,
            axis y line*=right,
            ylabel near ticks,
            xmin=50,
            xmax=180,
            ymin=0,
            ymax=9,
            xtick={50,75,100,125,150,175},
            minor x tick num=4,
            ytick={0,3,6,9},
            grid=major,
            xminorgrids=true,
            minor grid style={gray!20},
            line width=0.6pt,
            mark size=0.05pt,
        ]

        \addplot[mark=*]
        coordinates {
        (50,0.08)
        (55,0.09)
        (60,0.11)
        (65,0.15)
        (70,0.20)
        (75,0.26)
        (80,0.33)
        (85,0.42)
        (90,0.53)
        (95,0.67)
        (100,0.82)
        (105,0.98)
        (110,1.18)
        (115,1.45)
        (120,1.70)
        (125,2.00)
        (130,2.40)
        (135,2.80)
        (140,3.30)
        (145,3.80)
        (150,4.40)
        (155,5.10)
        (160,5.80)
        (165,6.60)
        (170,7.50)
        (175,8.40)
        };

        \addplot[
            black,
            thick,
            dashed
        ]
        coordinates {(100,0) (100,9)};

        \end{axis}
        \end{tikzpicture}
        \captionof{figure}{[$FOV_h^\circ$] vs. [RMSE $\times 0.01$]}
        \label{fig:dfov}
    \end{minipage}
\end{figure}

\begin{table}
    \centering
    \small
    \setlength{\tabcolsep}{5pt}
    \renewcommand{\arraystretch}{1.0}
    \resizebox{0.4\textwidth}{!}{
    \begin{tabular}{llccc}
        \toprule
        \multicolumn{5}{c}{\textbf{(a) Point cloud Segmentation on the S3DIS~\cite{armeni20163d}}} \\
        \midrule
        \textbf{Method} & Modality & Sup. & 3D mIoU(\%) & mAcc(\%)\\
        \midrule
        \multicolumn{5}{l}{\textit{closed vocabulary}} \\
        KPConv~\cite{thomas2019kpconv}            & 3D & {\color{red}\textbf{full}} & 67.1 & 72.8\\
        PointTransformer~\cite{zhao2021point}     & 3D & {\color{red}\textbf{full}} & 70.4 & 76.5\\
        PointTransformerV2~\cite{wu2022point}     & 3D & {\color{red}\textbf{full}} & 71.6 & 77.9\\
        PointTransformerV3~\cite{wu2024point}     & 3D & {\color{red}\textbf{full}} & 73.4 & 78.9\\
        Sonata~\cite{wu2025sonata}                & 3D & {\color{red}\textbf{full}} & 76.0 & 81.6\\
        Concerto~\cite{zhang2025concerto}         & 2D+3D & {\color{red}\textbf{full}} & \underline{77.4} & \underline{85.0}\\
        Mask3D-CSet~\cite{schult2023mask3d}$^\dagger$       & 3D & {\color{red}\textbf{full}} & 72.9 & 78.3\\
        \midrule
        \multicolumn{5}{l}{\textit{open vocabulary}} \\
        OpenMask3D~\cite{takmaz2023openmask3d}$^\dagger$ & 2D+3D & {\color{blue}\textbf{inst}} & 36.7 & 43.6\\
        SAM3~\cite{carion2025sam3segmentconcepts} (3D-lifted)$^\dagger$ & 2D+3D & {\color{green}\textbf{none}} & 44.7 & 57.2\\
        \textbf{JOPP-3D(f) (ours)}                & 2D+3D & {\color{green}\textbf{none}} & 59.4 & 70.1\\
        \textbf{JOPP-3D (ours)}                   & 2D+3D & {\color{blue}\textbf{inst}} & \textbf{80.9} & \textbf{87.0}\\
        \midrule
        \midrule
        \multicolumn{5}{c}{\textbf{(b) Point cloud Segmentation on the ToF-360~\cite{kanayama2025tof}}} \\
        \midrule
        \textbf{Method} & Modality & Pretrain & mIoU(\%) & mAcc(\%)\\
        \midrule
        \multicolumn{5}{l}{\textit{closed vocabulary}} \\
        PointTransformerV3~\cite{wu2024point}     & 3D & S3DIS & 18.6 & 25.1\\
        SFSS-MMSI~\cite{guttikonda2024SFSS}       & 2D+3D & S3DIS & \underline{23.2} & \underline{46.3}\\
        \midrule
        \multicolumn{5}{l}{\textit{open vocabulary}} \\
        \textbf{JOPP-3D(f) (ours)}                & 2D+3D & -- & \textbf{30.9} & \textbf{47.5}\\
        \bottomrule
    \end{tabular}}
    \caption{Point cloud semantic segmentation results on the S3DIS~\cite{armeni20163d} and the ToF-360~\cite{kanayama2025tof} datasets. The best performance for each category is shown in \textbf{bold}, and the second best is \underline{underlined}. Supervision: {\color{red}\textbf{full}}: full semantic supervision, {\color{blue}\textbf{inst}}: instance masks only, {\color{green}\textbf{none}}: no supervision. In (b), \emph{Pretrain} reports the dataset each method was trained on, as ToF-360 is testing-only. $^\dagger$: evaluated by us using the authors' released code/checkpoints.}
    \label{tab:semantic_miou}
    \vspace{-0.2cm}
\end{table}
\begin{table}
\centering
\tiny
\setlength{\tabcolsep}{5pt}
\renewcommand{\arraystretch}{1.0}
\resizebox{0.4\textwidth}{!}{
\begin{tabular}{llccc}
\toprule
\multicolumn{5}{c}{\textbf{(a) Panoramic Segmentation on Stanford-2D-3D-s~\cite{2017arXiv170201105A}}} \\
\midrule
\textbf{Method} & Modality & Views & Sup. & mIoU(\%)\\
\midrule
\multicolumn{5}{l}{\textit{Closed vocabulary}} \\
TangentImg~\cite{Eder_2020_CVPR}$^\star$ & RGB-D & single & {\color{red}\textbf{full}} & 52.5 \\
Trans4PASS+~\cite{zhang2024behind} in~\cite{sgat4pass} & RGB & single & {\color{red}\textbf{full}} & 53.6 \\
SGAT4PASS~\cite{sgat4pass} & RGB & single & {\color{red}\textbf{full}} & 56.4 \\
360BEV~\cite{teng2024360bev} & RGB-D & single & {\color{red}\textbf{full}} & 56.5 \\
PanoSAMic~\cite{chamseddine2026panosamic}$^\dagger$ & RGB-D & single & {\color{red}\textbf{full}} & \underline{61.7} \\
{Mask3D-CSet}~\cite{schult2023mask3d}$^\dagger$ & RGB-D & multi & {\color{red}\textbf{full}} & {60.8}\\
\midrule
\multicolumn{5}{l}{\textit{Open vocabulary}} \\
OpenMask3D~\cite{takmaz2023openmask3d}$^\dagger$ & RGB-D & multi & {\color{blue}\textbf{inst}} & 29.8 \\
OOOPS~\cite{zheng2024open}$^\star$ & RGB & single & {\color{blue}\textbf{inst}} & 41.1 \\
\textbf{JOPP-3D(f) (ours)} & RGB-D & multi & {\color{green}\textbf{none}} & 52.8 \\
SAM3~\cite{carion2025sam3segmentconcepts}$^\dagger$ & RGB & single & {\color{green}\textbf{none}} & 54.2 \\
\textbf{JOPP-3D (ours)} & RGB-D & multi & {\color{blue}\textbf{inst}} & \textbf{70.1} \\
\midrule
\midrule
\multicolumn{5}{c}{\textbf{(b) Panoramic Segmentation on the ToF-360~\cite{kanayama2025tof}}} \\
\midrule
\textbf{Method} & Modality & Views & Pretrain & mIoU(\%)\\
\midrule
\multicolumn{5}{l}{\textit{Closed vocabulary}} \\
HoHoNet~\cite{sun2021hohonet} & RGB & single & Stanford-2D-3D-s & {20.8} \\
SFSS-MMSI~\cite{guttikonda2024SFSS} & RGB-D & single & Stanford-2D-3D-s & 24.9 \\
PanoFormer~\cite{shen2022panoformer} & RGB & single & Stanford-2D-3D-s & \underline{28.1} \\
\midrule
\multicolumn{5}{l}{\textit{Open vocabulary}} \\
\textbf{JOPP-3D(f) (ours)} & RGB-D & single & -- & \textbf{30.7} \\
\bottomrule
\end{tabular}}
\caption{Panoramic semantic segmentation performance on the Stanford-2D-3D-s~\cite{2017arXiv170201105A} and the ToF-360~\cite{kanayama2025tof} datasets. \emph{Views} indicates whether a method segments a single panorama or aggregates several panoramas of the same scene; ToF-360 is a single-capture dataset, so every method there is single-view. All values follow the 1-fold protocol unless marked $^\star$: 3-fold is reported.
}
\label{tab:semantic_miou2}
\vspace{-0.2cm}
\end{table}
\begin{figure}[bp]
    \centering
    \captionsetup{type=figure}
        \includegraphics[width=0.75\linewidth]{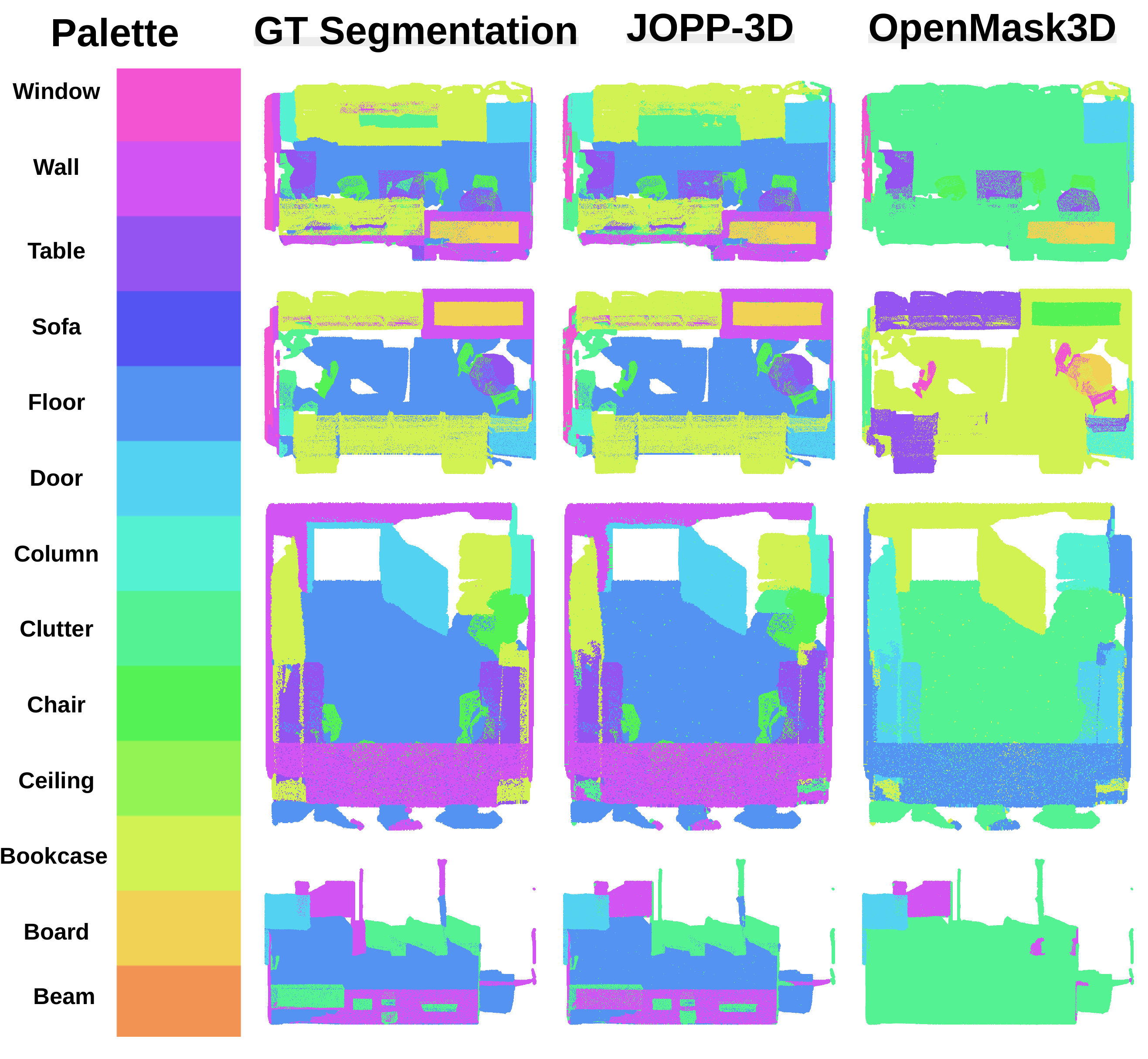}
        \captionof{figure}{S3DIS~\cite{armeni20163d} semantic 3D segmentation results comparison: We can clearly see JOPP-3D's ability to retrieve large structures like walls and floors. Whereas instance-centric formulations like OpenMask3D~\cite{takmaz2023openmask3d} confuse large structures with smaller structures like bookcases/chairs resulting in worse performance overall.}
    \label{fig:3d_op}
    \vspace{-0.2cm}
\end{figure}

\begin{table}
\centering
\tiny
\setlength{\tabcolsep}{5pt}
\renewcommand{\arraystretch}{0.3}
\resizebox{0.4\textwidth}{!}{
\begin{tabular}{lcccc}
\toprule
\multirow{2}{*}{\textbf{Method}} & \multirow{2}{*}{\textbf{Modality}} & \multirow{2}{*}{Base-train} & \multicolumn{2}{c}{AP$_{50}$ (\%)} \\ 
\cmidrule(lr){4-5}
 & & & B8/N4 & B6/N6 \\
\midrule
PLA~\cite{ding2023pla}          & 2D+3D      & \cmark &  8.6 &  9.8 \\
Lowis3D~\cite{ding2024lowis3d}  & 2D+3D      & \cmark & 13.8 & 15.8 \\
OpenMask3D~\cite{takmaz2023openmask3d}$^\dagger$ & 2D+3D & \xmark$^+$ & 23.7 & 29.8 \\
Open3DIS~\cite{nguyen2024open3dis} & 2D+3D     & \cmark & 26.3 & 29.0 \\
PoVo~\cite{povo2024} & 2D+3D             & \xmark & 29.1 & 33.4 \\
OpenIns3D~\cite{huang2024openins3d} & 3D   & \xmark$^+$ & {37.0} & 33.0 \\
\midrule
\textbf{JOPP-3D(f) (ours)} & 2D+3D   & \xmark & \underline{38.1} & \underline{41.9} \\
\textbf{JOPP-3D (ours)} & 2D+3D      & \xmark$^+$ & \textbf{51.7} & \textbf{60.7} \\
\bottomrule
\end{tabular}}
\caption{Open-vocabulary 3D {instance} segmentation on S3DIS Area-5: AP$_{50}$ on the
two base/novel splits. \emph{Base-train} marks
methods trained on the base split with semantic labels. $^+$: the class-agnostic 3D
proposal network is trained over all classes, so novel-class instances are seen
geometrically though never labelled.
}
\label{tab:s3dis_ap}
\end{table}

On the $F=20$ and $FOV_h=100^\circ$ choice: 
\noindent Faces of the regular icosahedron, is the densest \textit{uniform} subdivision of the sphere into planar faces.
Coarser polyhedra 
require larger per-face FOV, leading to more distortion. 
Table~\ref{tab:dfov} summarizes the resulting $FOV_h$ \& corresponding boundary distortion factor $D({FOV_h})=\sec^2(\frac{FOV_h}{2}).$
To the right in Fig.~\ref{fig:dfov}, we plot 5$^\circ$ $FOV_h$ intervals vs. projection error between the two unprojected point clouds of panoramic \& tangent depths.
The plot in Fig.~\ref{fig:dfov} rises sharply with $FOV_h$ after $100^\circ$ - combined with $D(FOV_h)$ behaviour from the table, we can geometrically ground our choices.
\subsection{Datasets}

The \textbf{Stanford-2D-3D-s}~\cite{2017arXiv170201105A} dataset contains high-resolution panoramic images paired with registered 3D point clouds, serving as a standard benchmark for both panoramic and 3D scene understanding. \textbf{S3DIS}~\cite{armeni20163d} is the 3D-only version of this dataset. We evaluate on the standard Fold-1 of Stanford-2D-3D-s (which is Area-5 of S3DIS) containing panoramic captures covering scenes such as \textit{offices}, \textit{hallways}, and \textit{conference rooms}. 

\noindent\textbf{ToF-360}~\cite{kanayama2025tof} is a testing-only dataset providing a complementary setup featuring synchronized Time-of-Flight~\cite{Horaud_2016} depth maps and panoramic RGB images along with corresponding semantic annotations. This dataset covers 4 areas which include \textit{offices}, \textit{hospital rooms}, and \textit{parking lot spaces}.

\subsection{Evaluation Metrics}
For panoramic semantic segmentation, we report two variants of mean Intersection-over-Union (mIoU): closed-vocabulary mIoU (the standard mIoU) and open-vocabulary mIoU (Open mIoU). 
The Open mIoU, following the evaluation proposed in \cite{zhou2025rethinking}, leverages semantic relationships from WordNet~\cite{wordnet} to consider conceptually similar class predictions (e.g., “\textit{sofa}” and “\textit{chair}”) as partial matches. 
To keep the comparison tables compact, Open mIoU is reported in the ablation of Table~\ref{tab:Ablations}, where every variant is open-vocabulary and the metric is directly comparable; the main tables report the standard mIoU, which all baselines share. 
For 3D semantic segmentation, we report the standard 3D mIoU and mean Accuracy (mAcc), which other semantic baselines evaluate on.
We report average precision (AP) separately in
Tab.~\ref{tab:s3dis_ap}, where we evaluate 3D instance segmentation.
\begin{figure}[tp]
    \centering
    \captionsetup{type=figure}
        \hspace{-0.9cm}\includegraphics[width=1.1\linewidth]{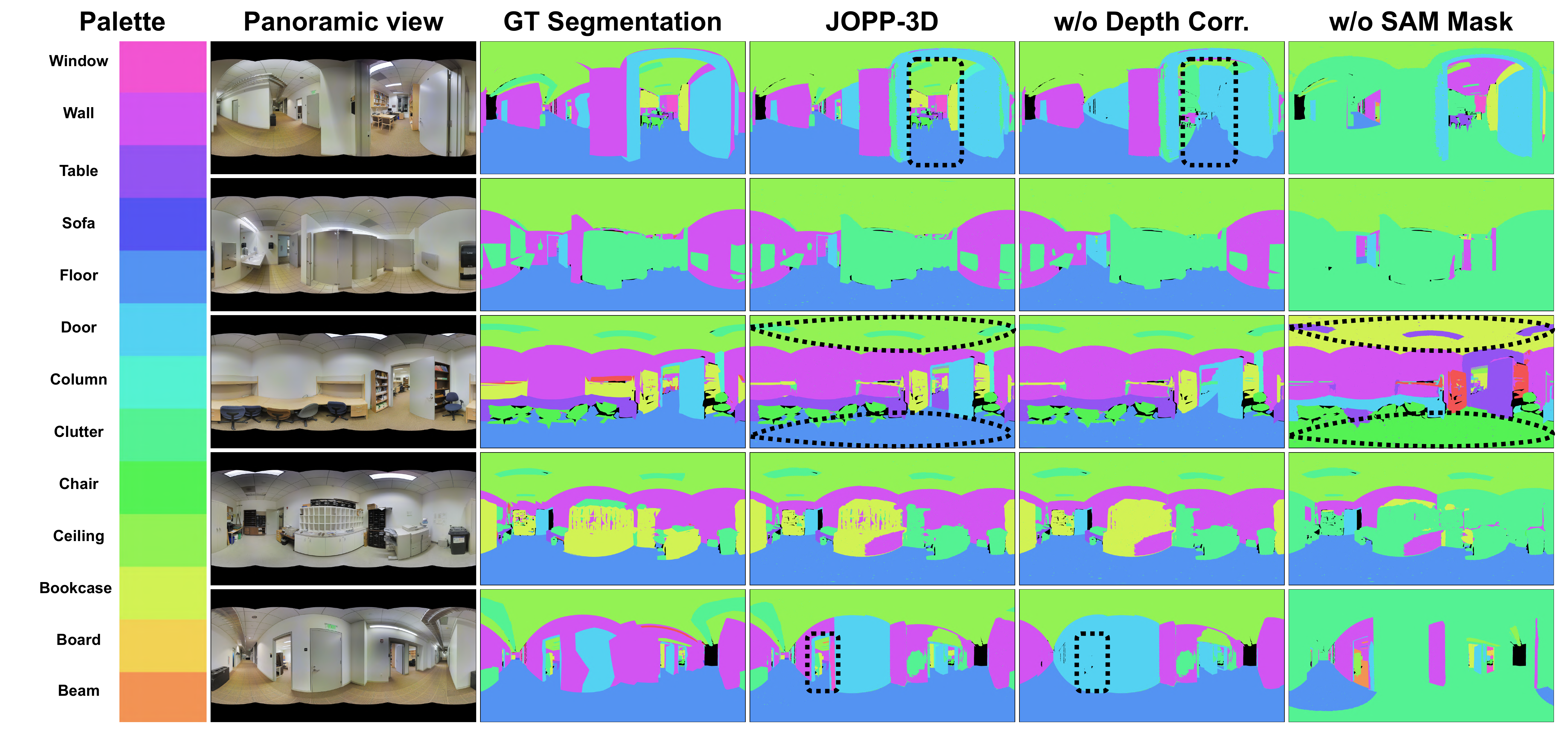}
        \captionof{figure}{Qualitative analysis: Each row represents one panoramic capture from a different scene in the Stanford-2D-3D-s~\cite{2017arXiv170201105A} dataset. We present the semantic segmentation result from various versions of JOPP-3D. Notable differences are highlighted via dotted curves. In the first and fifth row we can observe that our depth correspondence technique yields segmentations through doors into nearby scenes. In the third row we can see a clear problem of not masking the SAM~\cite{segmentAnything} crops - \textit{floor} and \textit{ceiling} cannot be segmented. Further, querying for \textit{chair} also selects \textit{floor}. Note: The last column has a lot of \textit{clutter} palette. Reason: For \textit{clutter}, we assign all the 3D points that could not be queried using the other 12 classes, and as expected, w/o SAM Mask ablation suffers the worst due to bad CLIP~\cite{radford2021learning} embedding alignment.}
    \label{fig:ablation}
    \vspace{-0.7cm}
\end{figure}

\begin{table*}
  \renewcommand{\arraystretch}{1.0}
  \small
  \resizebox{1.0\textwidth}{!}{\begin{tabularx}{1.2\textwidth}{lc*{13}{X}}
    \hline
    \rotatebox{65}{{\renewcommand{\arraystretch}{0.8}\begin{tabular}{l}\textbf{Method}\\\textbf{Variant}\end{tabular}}}
    & \rotatebox{65}{{\renewcommand{\arraystretch}{0.8}\begin{tabular}{l}Open\\mIoU (\%)\end{tabular}}}
    & \rotatebox{65}{Beam} 
    & \rotatebox{65}{Board} 
    & \rotatebox{65}{Bookcase} 
    & \rotatebox{65}{Ceiling} 
    & \rotatebox{65}{Chair} 
    & \rotatebox{65}{Clutter} 
    & \rotatebox{65}{Column} 
    & \rotatebox{65}{Door} 
    & \rotatebox{65}{Floor} 
    & \rotatebox{65}{Sofa} 
    & \rotatebox{65}{Table} 
    & \rotatebox{65}{Wall} 
    & \rotatebox{65}{Window}
\\
    \hline
    \hline
    w/o SAM Mask  & 33.6 & 12.1 & 56.0 & 20.2 & 34.2 & 52.7 & 40.9 & 19.6 & 32.4 & 30.2 & 53.3 & 30.8 & 29.0 & 25.6\\
    w/o Tgt Decomp.                         & 41.4 & 22.4 & 55.8 & 38.4 & 41.4 & 60.3 & 28.2 & 31.4 & 40.1 & 33.1 & 16.6 & 43.5 & 36.8 & 48.7\\
    JOPP-3D(f)          & 59.9 & 22.0 & 87.6 & 54.5 & 68.5 & 53.1 & 40.9 & 28.2 & 52.5 & 75.9 & 74.0 & 53.5 & 85.9 & 60.5\\
    w/o Depth Corr.                     & {67.0} & 15.1 & 80.3 & 62.8 & 83.2 & 74.7 & 54.2 & 66.0 & 75.0 & 90.7 & 62.6 & 75.9 & 79.8 & 50.2\\
    \hline
    \textbf{JOPP-3D}        & \textbf{74.6} & \textbf{35.8} & \textbf{86.8} & \textbf{68.2} & \textbf{88.5} & \textbf{85.1} & \textbf{59.0} & \textbf{73.0} & \textbf{82.4} & \textbf{95.2} & \textbf{81.1} & \textbf{81.8} & \textbf{86.6} & \textbf{68.6}\\
    \hline
  \end{tabularx}}
 \caption{Ablation results on the Stanford-2D-3D-s~\cite{2017arXiv170201105A}. Ablation studies are done by taking out one component at a time.}
  \label{tab:Ablations}
  \vspace{-0.2cm}
\end{table*}
\subsection{Quantitative Results}

Table~\ref{tab:semantic_miou}(a) reports results on the S3DIS Area-5 for 3D semantic segmentation, where JOPP-3D reaches \textbf{80.9\%} mIoU and \textbf{87.0\%} mAcc. We note that the comparison against 3D-only methods: KPConv~\cite{thomas2019kpconv} through Sonata~\cite{wu2025sonata} is not direct. 
Therefore, comparison at matched input is against Concerto~\cite{zhang2025concerto}, the strongest supervised entry in the table, which JOPP-3D exceeds while requiring no semantic supervision. Similarly, Table~\ref{tab:semantic_miou2}(a) presents the comparison of JOPP-3D against recent panoramic semantic segmentation methods on the Stanford-2D-3D-s, where we obtain \textbf{70.1\%} mIoU. Here the gain over the supervised single-panorama methods comes in part from aggregating several captures of a scene, which is an advantage of operating in 3D. 

Table~\ref{tab:semantic_miou}(b), and Table~\ref{tab:semantic_miou2}(b) presents JOPP-3D(f) results on the ToF-360 evaluation dataset. ToF-360 is a challenging benchmark for zero-shot model evaluation, where once again, our variant shows clear improvements. 
To quantify Mask3D~\cite{schult2023mask3d} dependence, we report a controlled baseline, \emph{Mask3D-CSet}, using the same Mask3D instance proposals + a supervised semantics trained head in Tables~\ref{tab:semantic_miou}, \ref{tab:semantic_miou2}. Despite full closed-set supervision, Mask3D-CSet trails JOPP-3D on the same 13 classes - direct evidence that our semantic alignment, not Mask3D by itself, drives the headline result. We also lift the SAM3~\cite{carion2025sam3segmentconcepts} predictions to 3D using our JOPP-3D(f) lifting strategy. While SAM3 has slightly better results in pano-domain, JOPP-3D(f) clearly overperforms SAM3 in the 3D domain. These two sets of experiments demonstrate the effectiveness of our joint approach - 3D-specific (Mask3D) or image-specific (SAM3) approaches cannot directly perform better in their domain counterparts.

Notably, OpenMask3D~\cite{takmaz2023openmask3d} in both Table~\ref{tab:semantic_miou},~\ref{tab:semantic_miou2} struggles with semantic segmentation due to its instance-centric design choices and its dependence on sequenced RGB-D frames 
(see Fig.~\ref{fig:3d_op}). Beyond semantic segmentation, Table~\ref{tab:s3dis_ap} scores the same predictions as 3D \emph{instance} segmentation under the base/novel protocol of PLA~\cite{ding2023pla}, reporting AP$_{50}$ on the novel classes against a broad set of recent open-vocabulary methods. JOPP-3D(f) already surpasses all of them, while performing no 3D training of its own. JOPP-3D is further ahead by a wide margin, which follows directly due to the higher quality of its instance proposals than JOPP-3D(f).

\subsection{Inference Cost}
\label{sec:inference_cost}
JOPP-3D pipeline's runtime cost separates into a
one-time stage and a per-query stage. Tangential decomposition, 3D reconstruction and
language-guided semantic alignment average $4.8$\,min per $4096\times2048$ capture, but are
performed \emph{once} per scene and are never repeated: the per-mask embeddings are cached,
so the vocabulary can be extended afterwards at no additional alignment cost. Every
subsequent open-vocabulary query then costs $1.7$\,s to obtain the 3D semantic segmentation,
plus one rasterization of the 3D prediction back onto the sphere to produce the panoramic
output, as reported in Tab.~\ref{tab:inference_cost}. At our native $4096\times2048$ output
the rasterization takes $19.7$\,s, while at the $1024\times512$ at which other panoramic baselines
report, it takes $1.5$\,s. (further elaborated in supplementary sec.D)

\subsection{Ablation Study}
\label{sec:ablations}

To better understand the contributions of each component in our framework, we conduct an ablation study on several variants of our method, 
as shown in Table~\ref{tab:Ablations}
and Figure~\ref{fig:ablation}.

\noindent\textbf{Effect of SAM Masking:} The variant \textit{w/o SAM Mask} removes the 2D masking step from the instance crops. Without this, CLIP embeddings are computed directly on these instance crops, which often contain multiple objects within large instances such as \textit{floor}, \textit{ceiling}, and \textit{walls}. This leads to poor class distinction, resulting in the lowest performance.

\noindent\textbf{Tangential Decomposition:} Using full panoramic images instead of tangentially decomposed perspective views (\textit{w/o Tgt Decomp.}) also degrades performance due to inherent distortions in panoramas, hindering CLIP embeddings.

\noindent\textbf{JOPP-3D(f) vs. JOPP-3D:} Replacing Mask3D~\cite{schult2023mask3d} with our SAM3D~\cite{sam3d} adaptation yields moderate gains over the previous ablations, showing the limitations of JOPP-3D - the quality of instance proposals affects performance the most.

\noindent\textbf{Depth Correspondence:} Finally, removing depth-based correspondence (\textit{w/o Depth Corr.}) 
leads to lack of precise 3D-to-pano semantic association resulting in a 
consistent loss of performance across all categories compared to our full model, JOPP-3D.
Overall, the ablation study confirms that each of our components contribute substantially.

\begin{table}[t]
\centering
\small
\setlength{\tabcolsep}{5pt}
\renewcommand{\arraystretch}{1.0}
\resizebox{0.5\textwidth}{!}{
\begin{tabular}{lccc}
\toprule
\multicolumn{4}{c}{\textbf{Per-query inference cost}} \\
\midrule
\textbf{Output resolution} & 3D query(s) & Rasterization(s) & Total(s)\\
\midrule
$4096\times2048$ (ours)            & 1.7 & 19.7 & 21.4 \\
$2048\times1024$                   & 1.7 & 5.2  & 6.9 \\
$1024\times512$ (baseline setting) & 1.7 & 1.5  & \textbf{3.2} \\
\bottomrule
\end{tabular}}
\caption{Cost of a single open-vocabulary query and rasterization, once a
scene has been processed.(on one NVIDIA RTX A6000)}
\label{tab:inference_cost}
\vspace{-0.2cm}
\end{table}
\section{Conclusion}

We introduced JOPP-3D, a 
joint open-vocabulary segmentation 
framework that unifies point cloud and panoramic scene understanding by exploring the equivalence between the two modalities.
The proposed model pipeline
consisting of tangential decomposition,
VLM-based semantic alignment, and depth-guided 3D-to-pano correspondence - effectively bridges panoramic imagery with 3D point clouds. 

Through extensive experiments on the Stanford-2D-3D-s and ToF-360 datasets, we demonstrated that JOPP-3D achieves superior performance compared to existing panoramic-only or point cloud-only semantic segmentation approaches, while uniquely performing both tasks jointly. 
Our ablation study further highlights the importance of each design choice, while also showing the limitations: the dependence on the quality of instance proposals. 

Therefore, a clear future work is to find a more reliable 2D$\to$3D instance-lifting strategy, that can allow a higher JOPP-3D(f) performance, and move towards complete zero-shot inference. Moreover, JOPP-3D's pipeline restricts the current system to offline processing. Therefore, supporting incremental updates as new panoramas arrive is another future direction.
Overall, JOPP-3D represents an important step towards 
scene understanding that connects open-vocabulary reasoning between panoramic and 3D spaces. 
\section*{Acknowledgements}
This research was funded by the European Union as part of the project: ShieldBOT (Grant Agreement 101235093).
{
    \small
    \bibliographystyle{ieeenat_fullname}
    \bibliography{main}
}
\newpage
\setcounter{section}{0}
\renewcommand\thesection{\Alph{section}}
\renewcommand\thefigure{\thesection.1}
\section{Implementation details for SAM3D}
The original SAM3D~\cite{sam3d} implementation assumes an input of RGB-D image sequences, where continuous merging is performed by pairing two adjacent frames with sufficient overlap. In contrast, our input consists of discretely captured panoramic RGB-D images, which do not guarantee strong overlap between any two panoramas. To address this, we treat all tangential perspectives generated from a single panorama as an adjacent group and perform sequential merging within this group. We refer to this process as local merging, while merging across panoramas is referred to as global merging. 

\begin{figure}[htbp]
    \centering
    \textbf{Colored Pointcloud}
    \begin{subfigure}{1\linewidth}
        \begin{subfigure}{0.495\linewidth}
            \begin{subfigure}{\linewidth}
                \centering
                \includegraphics[width=\linewidth]{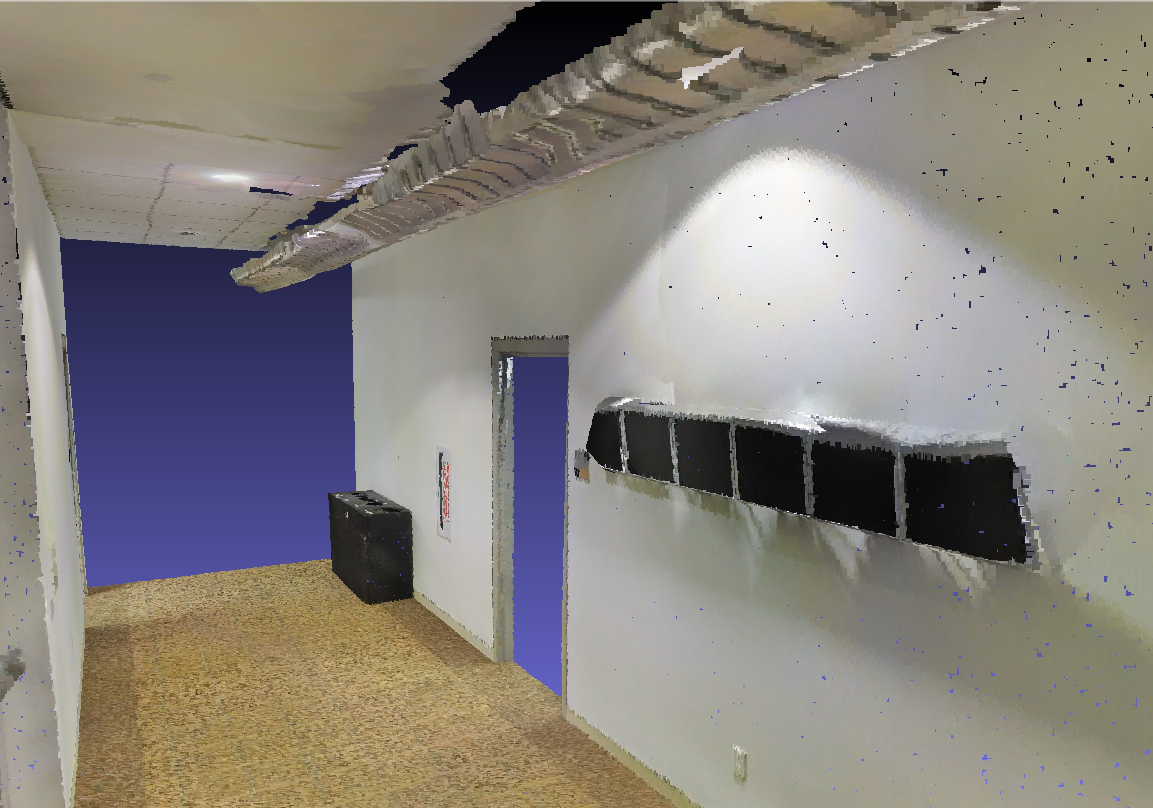}
            \end{subfigure}
        \end{subfigure}
        \begin{subfigure}{0.495\linewidth}
            \begin{subfigure}{\linewidth}
                \centering
                \includegraphics[width=\linewidth]{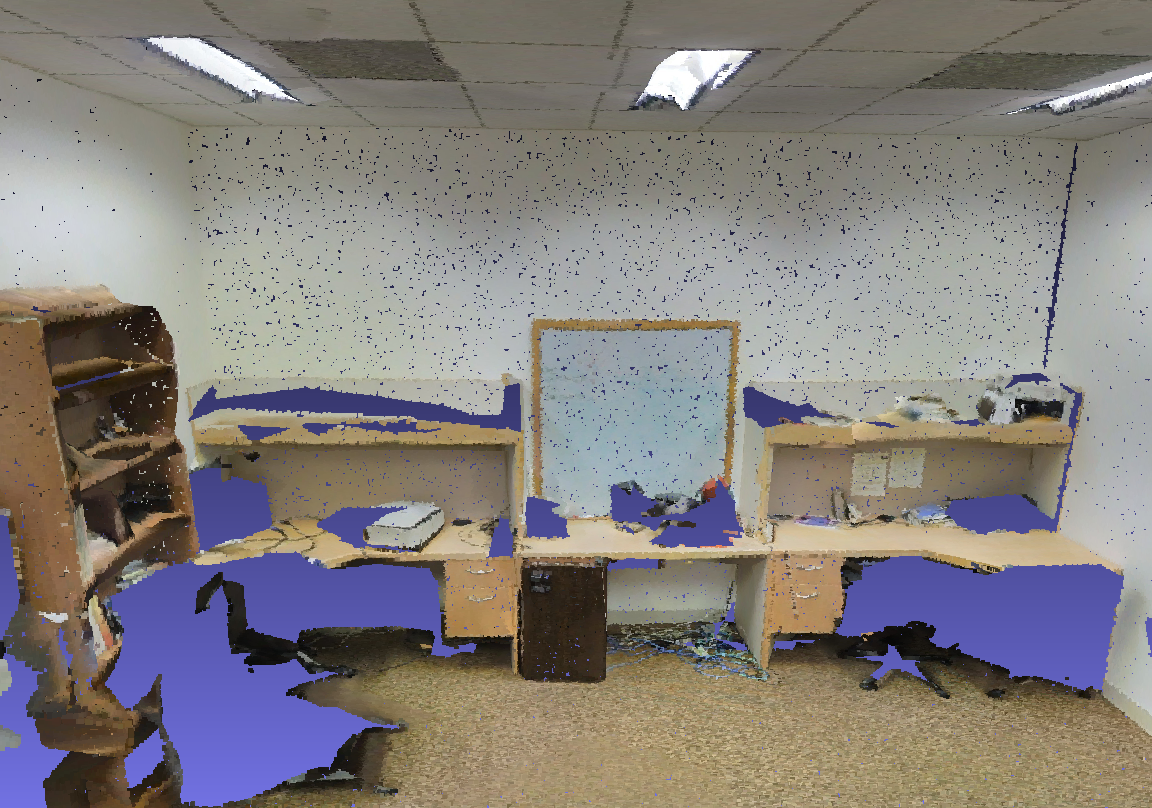}
            \end{subfigure}
        \end{subfigure}
    \end{subfigure}
    \textbf{SAM3D (Original)}
    \begin{subfigure}{1\linewidth}
        \begin{subfigure}{0.495\linewidth}
            \begin{subfigure}{1\linewidth}
                \includegraphics[width=1\linewidth]{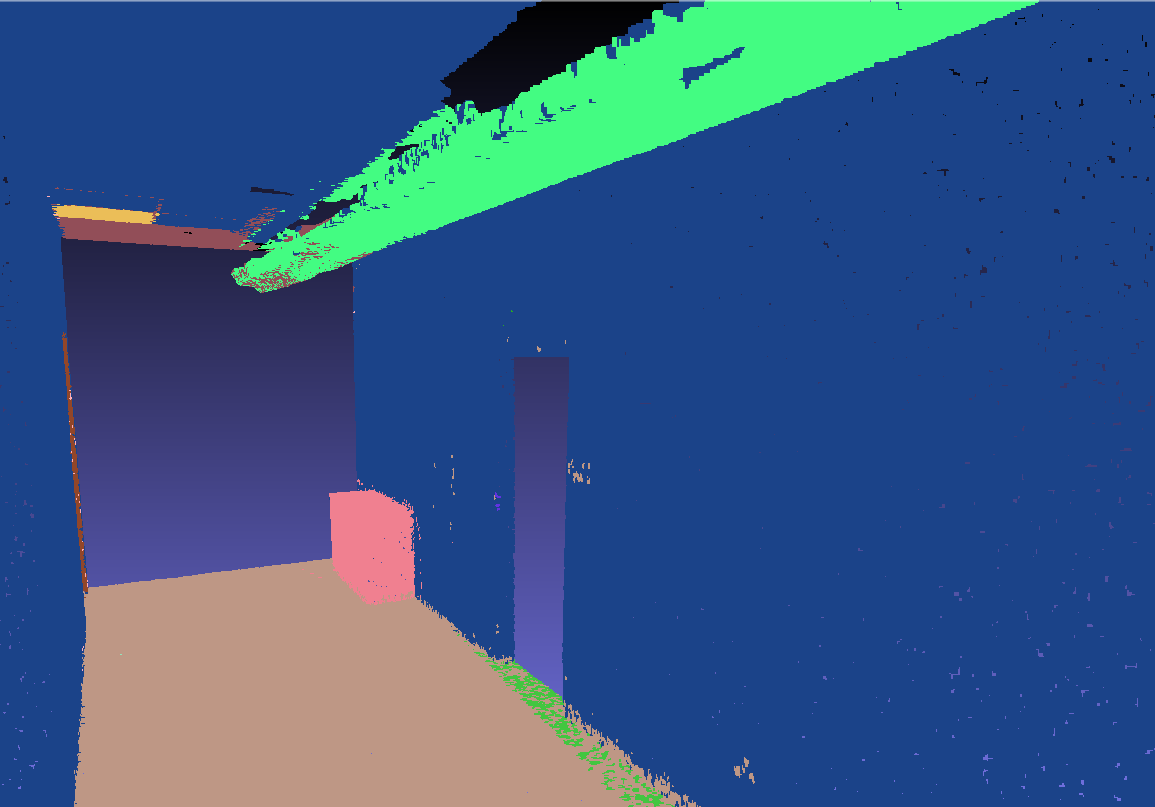}
            \end{subfigure}
        \end{subfigure}
        \begin{subfigure}{0.495\linewidth}
            \begin{subfigure}{1\linewidth}
                \includegraphics[width=1\linewidth]{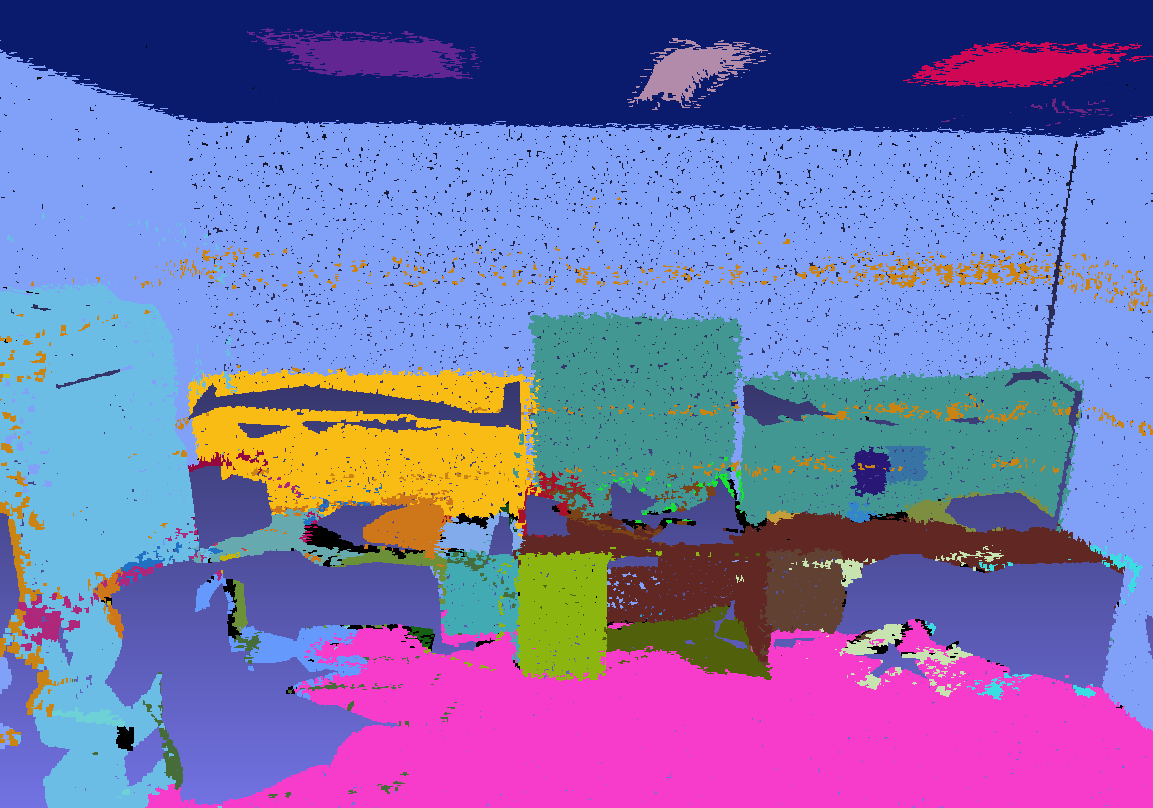}
            \end{subfigure}
        \end{subfigure}
    \end{subfigure}
    \textbf{SAM3D (Ours)}
    \begin{subfigure}{1\linewidth}
        \begin{subfigure}{0.495\linewidth}
            \begin{subfigure}{1\linewidth}
                \includegraphics[width=1\linewidth]{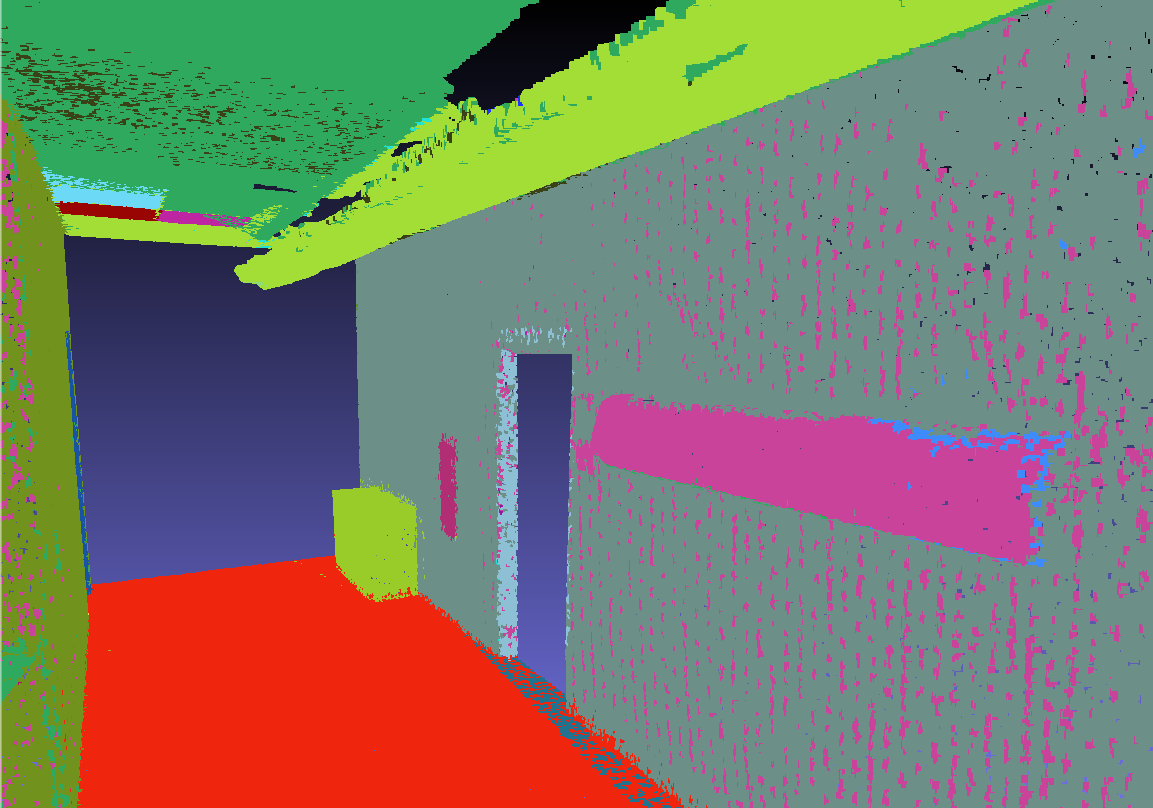}
            \end{subfigure}
        \end{subfigure}
        \begin{subfigure}{0.495\linewidth}
            \begin{subfigure}{1\linewidth}
                \includegraphics[width=1\linewidth]{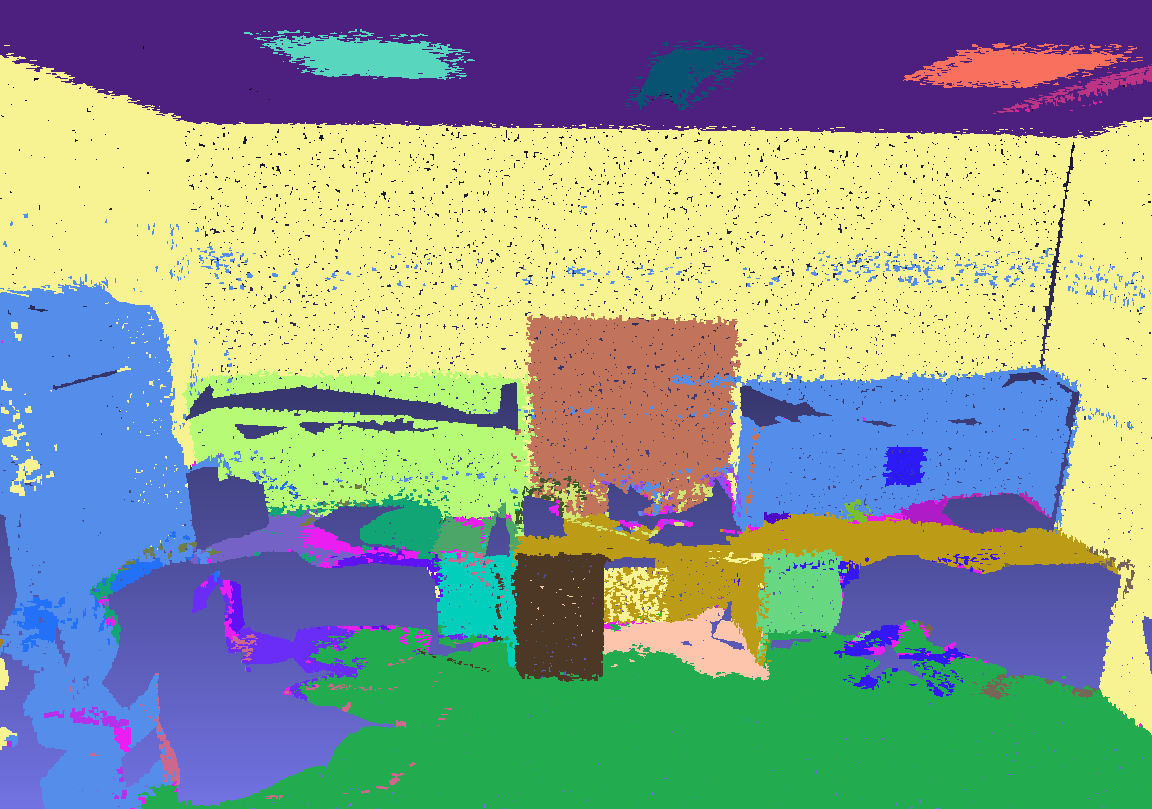}
            \end{subfigure}
        \end{subfigure}
    \end{subfigure}
    \caption{Instance segmentation comparison across two different scenes. Rows: Top shows the colored point cloud input; middle shows results from the original SAM3D; bottom shows results from our improved SAM3D with tangent-decomposition-based local and global merging. Columns: Left column corresponds to a hallway scene, and right column to an office scene. In the hallway scene, the original SAM3D merges the ceiling and wall into a single instance, whereas our method achieves clear separation. In the office scene, the original SAM3D merges the whiteboard and desk, while our approach successfully distinguishes them as separate instances.}
    \label{fig:sam3d_comparison}
\end{figure}

Both local and global merging follow an overlap-based criterion: two instance masks are merged if their intersection-over-union (IoU) exceeds a predefined threshold. As merged instances grow larger, a higher threshold is required to maintain proper instance separation. In our experiments, we set a relatively low threshold for local merging $(0.2)$ to allow fine-grained merging within a panorama, and a higher threshold for global merging $(0.2-0.4)$ to avoid over-merging across panoramas. In contrast, the original SAM3D continuously merges adjacent frames using the same IoU threshold regardless of the merging stage (see Fig.~\ref{fig:sam3d_comparison}).

\section{Rare-class retrieval}
\setcounter{figure}{0}
\renewcommand\thefigure{\thesection.\arabic{figure}}
A key challenge during the evaluation of JOPP-3D arises from a dataset’s reliance on generic labels due to limitations in annotating capacity. 
For example, in the Stanford-2D-3D-s~\cite{2017arXiv170201105A} dataset, a large variety of semantically meaningful objects -- including wall decorations (e.g. clocks, posters), dustbins, and other small furniture are all collapsed into a single “clutter” category. This poses a disadvantage for an open-vocabulary approach such as JOPP-3D, whose strength lies in retrieving semantically coherent concepts using natural-language queries. In contrast, fully supervised methods can simply be trained to deal with “clutter” from large quantities of annotated examples, without having to distinguish the underlying object types.

\begin{figure}
    \centering
    \captionsetup{type=figure}
        \includegraphics[width=1.0\linewidth]{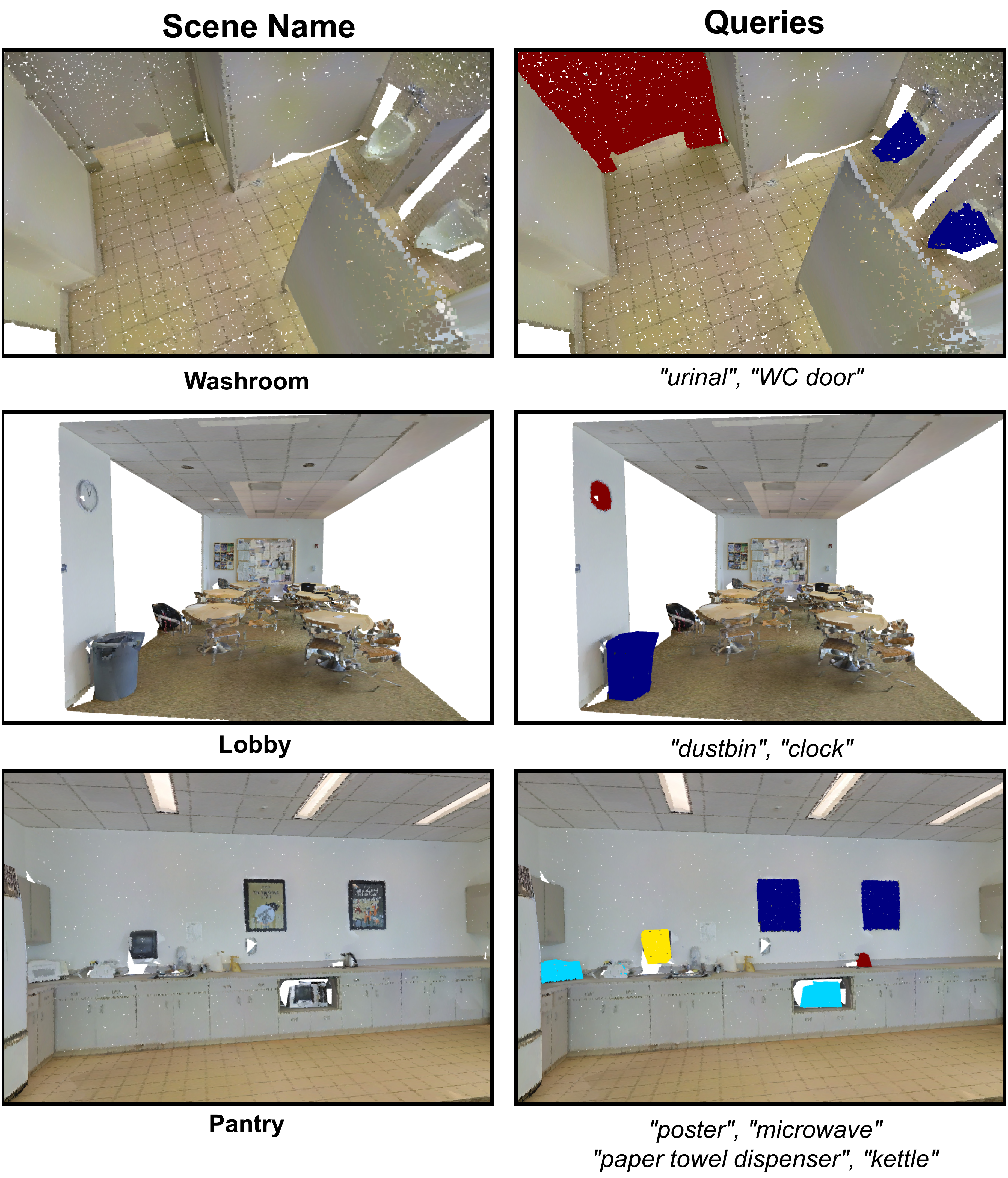}
        \captionof{figure}{First Column: Various 3D scenes, Second Column: The corresponding semantic masks from queries. These objects were marked under “clutter” category in ground-truth semantic annotations, however, JOPP-3D can easily retrieve them.}
    \label{fig:clutter}
\end{figure}
\renewcommand\thetable{\thesection.1}

To better highlight this limitation of the dataset and the advantage of open-vocabulary retrieval, we present additional qualitative results where we directly query fine-grained object categories inside 3D scenes. As shown in Fig.~\ref{fig:clutter}, these concepts can be reliably retrieved using our language-guided 3D semantic alignment mechanism, even though all of these items are collapsed into the same clutter label in the ground-truth annotations. While this limits our quantitative scores under the dataset's closed-set metric, it emphasizes the practical utility of JOPP-3D in real-world environments where semantically rich understanding is required beyond fixed label vocabularies.
\begin{table}[bp]
\renewcommand{\arraystretch}{1.0}
  \resizebox{1.0\linewidth}{!}{\begin{tabular}{l|c|c|c|c|c|c|c}
\toprule
Query & dustbin & printer & backpack & water cooler & monitor & microwave & cardboard box \\
\midrule
$\text{Hit@}10_{50}$ (\%) & 86.4 & 84.8 & 84.8 & 83.3 & 80.3 & 78.8 & 75.8 \\
Unique clutter     & 43   & 51   & 25   & 40   & 45   & 33   & 41   \\
\bottomrule
\end{tabular}}
\caption{The top rare-classes able to be retrieved by JOPP-3D. All these rare-class instances are grouped under "clutter".}
  \label{tab:rareclass}
\end{table}
 In Table~\ref{tab:rareclass}, we provide rare-class retrieval on a total 922 clutter instances in S3DIS~\cite{armeni20163d}. We provide two rare-retrieval metrics:
 \begin{itemize}
     \item $\text{Hit@}10_{50}$: fraction of rare-class queries for which at least one of the top-10 retrieved instances has  3D IoU $\ge 50\%$ with the GT-clutter region. 
     \item \emph{Unique clutter} counts the distinct GT-clutter instances localized by each query as its highest-similarity retrieval.
 \end{itemize}
Figs.~\ref{fig:rare2a} and~\ref{fig:rare2b} give eight further examples across offices, a
hallway and a washroom, each retrieved by a single natural-language query and each
annotated only as \textit{clutter} in the ground truth.

\section{Synergy between the 3D pipeline \& panoramas} 
In our method, panoramas provide us with the tangential decomposition for visual feature extraction and enable language-guided semantic alignment. Conversely, the 3D pipeline also synergises back, in the following ways:
\begin{itemize}
    \item \emph{Full-resolution panoramic prediction:} Current panoramic-segmentation methods struggle at native $4096\times2048$ resolution \& report results at $1024\times512$, a 16X lower resolution. By operating in 3D 
with our semantic label propagation, we produce predictions at native panorama resolution efficiently.
\item \emph{Cross-panorama consistency:} Independent inference on different panoramas yields inconsistent labels for the same instance viewed from different viewpoints; our joint pipeline solves this by maintaining a single canonical 3D prediction.
\end{itemize}

\section{Computational Efficiency}
\setcounter{figure}{0}
\renewcommand\thefigure{\thesection.\arabic{figure}}
Quantifying computational efficiency for JOPP-3D presents a unique scenario, as our method is inference-only given the instance proposals. In contrast, most baselines in both panoramic and 3D semantic segmentation require supervised training on large-scale datasets using substantial GPU resources. Nevertheless, following standard compute reporting guidelines, we provide an approximate analysis of inference-time compute usage and efficiency. 

The majority of JOPP-3D’s compute cost arises from 
the depth-correspondence mechanism used for consistent 3D-to-panoramic semantic transfer.
We implement $\delta$ in Eq.~15 with an equivalent KD-tree-based formulation that thresholds the number of overlapping 3D points, $N$($\sim$200K) between adjacent panoramas, $P_{\text{adj}}$($\sim$10) - Reduces complexity from $O(P_{\text{adj}}^2 \cdot H\cdot W)$ to a single $O(H\cdot W \cdot \log N)$ KD-tree query, while remaining invariant to local depth scale.
Fig.~\ref{fig:complexity} illustrates the two formulations and the resulting cost. The
saving is structural rather than constant-factor: the pairwise formulation compares every
pair of adjacent panoramas and so grows quadratically in $P_{\text{adj}}$, whereas merging
the overlapping points into a single tree makes the cost independent of how many panoramas
are adjacent. The two are comparable for a handful of captures and diverge from there, which
matters because dense captures are exactly the regime panoramic scanning produces.
\definecolor{cxNaive}{HTML}{D55E00}
\definecolor{cxOurs}{HTML}{0072B2}
\definecolor{cxInk}{HTML}{3F3F46}
\definecolor{cxMuted}{HTML}{A1A1AA}

\begin{figure}[t]
\centering
\begin{tikzpicture}[scale=0.82, transform shape,
    pano/.style={circle, draw=cxInk, fill=white, inner sep=1.3pt, font=\tiny},
    tree/.style={rounded corners=2pt, draw=cxOurs, fill=cxOurs!8, inner sep=3pt,
                 font=\tiny, align=center}]

  \begin{scope}[shift={(0,0)}]
    \node[font=\scriptsize\bfseries, text=cxInk] at (0,1.5) {Pairwise};
    \foreach \i in {0,...,4} {
      \node[pano] (n\i) at ({90+\i*72}:0.95) {$P_{\i}$};
    }
    \foreach \a/\b in {0/1,0/2,0/3,0/4,1/2,1/3,1/4,2/3,2/4,3/4} {
      \draw[cxNaive, opacity=0.55, line width=0.4pt] (n\a) -- (n\b);
    }
    \node[font=\tiny, text=cxNaive] at (0,-1.5)
         {$\mathcal{O}(P_{\mathrm{adj}}^{2}\!\cdot\! H W)$};
  \end{scope}

  \begin{scope}[shift={(4.1,0)}]
    \node[font=\scriptsize\bfseries, text=cxInk] at (0,1.5) {KD-tree (ours)};
    \node[tree] (kd) at (0,0) {KD-tree\\$N\!\approx\!200$K};
    \foreach \i in {0,...,4} {
      \node[pano] (m\i) at ({90+\i*72}:0.95) {$P_{\i}$};
      \draw[cxOurs, ->, line width=0.5pt] (m\i) -- (kd);
    }
    \node[font=\tiny, text=cxOurs] at (0,-1.5)
         {$\mathcal{O}(H W\!\cdot\!\log N)$};
  \end{scope}
\end{tikzpicture}

\vspace{2pt}

\begin{tikzpicture}
\hspace{-0.8cm}
\begin{axis}[
  width=0.92\linewidth, height=3.9cm,
  scale only axis,
  xmin=1, xmax=25, ymin=0.04, ymax=60,
  ymode=log, log basis y=10,
  xlabel={\small Adjacent panoramas $P_{\mathrm{adj}}$},
  ylabel={\small Relative cost},
  xlabel style={text=cxInk, font=\scriptsize},
  ylabel style={text=cxInk, font=\scriptsize},
  tick label style={font=\tiny, text=cxInk},
  axis line style={draw=cxMuted, line width=0.4pt},
  tick style={draw=cxMuted},
  grid=major,
  grid style={draw=cxMuted!35, line width=0.3pt},
  legend style={font=\tiny, draw=cxMuted!60, fill=white, at={(0.02,0.97)},
                anchor=north west, row sep=-1pt},
  legend cell align=left,
]
  \addplot[cxNaive, line width=1.2pt, domain=1:25, samples=60, smooth]
    {x^2/17.6};
  \addlegendentry{Pairwise $\mathcal{O}(P_{\mathrm{adj}}^{2} H W)$}
  \addplot[cxOurs, line width=1.2pt, domain=1:25, samples=2] {1};
  \addlegendentry{KD-tree $\mathcal{O}(H W \log N)$}

  \addplot[only marks, mark=*, mark size=1.7pt, cxOurs] coordinates {(10,1)};
  \node[font=\tiny, text=cxOurs, anchor=south] at (axis cs:10,1.25) {ours, $5.7\times$ cheaper};
  \addplot[only marks, mark=*, mark size=1.7pt, cxNaive] coordinates {(10,5.68)};
\end{axis}
\end{tikzpicture}

\caption{Cost of the depth-correspondence step $\delta$ (Eq.~15). \textbf{Top:} comparing
every pair of adjacent panoramas costs $\mathcal{O}(P_{\mathrm{adj}}^{2} H W)$, whereas
merging their overlapping points into one KD-tree and issuing a single nearest-neighbour
query per pixel costs $\mathcal{O}(H W \log N)$ and is \emph{independent} of how many
panoramas are adjacent. \textbf{Bottom:} cost relative to our formulation, for
$H W = 4096\times2048$ and $N\approx200$K. The two cross at $P_{\mathrm{adj}}\approx4$; at
our operating point of $P_{\mathrm{adj}}\approx10$ the pairwise formulation is $5.7\times$
more expensive, and the gap grows quadratically thereafter.}
\label{fig:complexity}
\end{figure}

The split of this cost into a one-time stage and a per-query stage is reported in the
\emph{Inference Cost} subsection of the main paper. 
Although compute-efficiency comparisons are not directly applicable to training-free methods, we contextualize our results against the open-vocabulary baseline, Open Panoramic Segmentation~\cite{zheng2024open}, which obtains 41.1 mIoU. JOPP-3D achieves 70.1 mIoU -- an improvement of more than 70\%. When normalizing compute relative to performance gain, we achieve  
$\approx$ \textbf{0.54 GPU+CPU hours per percentage point}. This relatively low cost highlights the computational efficiency of JOPP-3D, especially given its ability to jointly perform open-vocabulary segmentation in both panoramic and 3D domains.

\noindent\textbf{Same-hardware comparison against direct panoramic alignment.}
All experiments are conducted on a single workstation equipped with
an Intel Xeon 12-core CPU with 64\,GB RAM and a single NVIDIA RTX A6000 GPU with 48\,GB
memory. The alternative to tangential decomposition is to align features on the panoramas
themselves, which is what OpenMask3D~\cite{takmaz2023openmask3d} does when applied to this input. 

We compare the two, over the same scenes, the same Mask3D
proposals, so that the only difference is the alignment strategy.

Two properties separate the two. First, direct panoramic alignment invokes the
depth-propagation step of Fig.~\ref{fig:complexity} \emph{twice} per capture, once to
associate 3D proposals with panorama pixels during alignment and once to produce the
panoramic output, whereas tangential decomposition needs it only for the output: our
alignment reads tangential views, which carry their own poses and need no propagation.
Second, and more consequential, direct alignment must score every proposal against every
panorama, so its cost carries a proposals $\times$ captures interaction, while ours visits
the $K$ most visible tangential views per proposal and is therefore \emph{constant} in the
number of captures. The two are comparable on small scenes and separate as scenes grow,
which is the regime that matters. 

Fig.~\ref{fig:runtime_scaling} reports the measurement over eight Area-5 scenes spanning $1$
to $25$ captures, and Tab.~\ref{tab:runtime_scaling} lists the underlying timings. Fitting
cost per proposal against capture count shows us that JOPP-3D becomes
$2.0\times$ faster per capture as the number of panoramic captures to process increases. 

\providecommand{\rtJO}{}\definecolor{rtJO}{HTML}{0072B2}
\providecommand{\rtOM}{}\definecolor{rtOM}{HTML}{D55E00}
\providecommand{\rtInk}{}\definecolor{rtInk}{HTML}{3F3F46}
\providecommand{\rtMuted}{}\definecolor{rtMuted}{HTML}{A1A1AA}

\begin{figure}[t]
\centering
\begin{tikzpicture}
\begin{axis}[
  width=0.86\linewidth, height=4.3cm, scale only axis,
  xmin=0, xmax=27, ymin=0, ymax=19,
  xlabel={\small Panoramic captures in the scene},
  ylabel={\small Alignment time / proposal (s)},
  xlabel style={text=rtInk, font=\scriptsize},
  ylabel style={text=rtInk, font=\scriptsize},
  tick label style={font=\tiny, text=rtInk},
  axis line style={draw=rtMuted, line width=0.4pt},
  tick style={draw=rtMuted},
  grid=major, grid style={draw=rtMuted!35, line width=0.3pt},
  legend style={font=\tiny, draw=rtMuted!60, fill=white, at={(0.02,0.97)},
                anchor=north west, row sep=-1pt},
  legend cell align=left,
]
  \addplot[rtOM, line width=1.2pt, domain=0:27, samples=2] {0.641*x + 1.08};
  \addlegendentry{OpenMask3D, direct panoramic}
  \addplot[rtJO, line width=1.2pt, domain=0:27, samples=2] {0.314*x + 3.87};
  \addlegendentry{\textbf{JOPP-3D}, tangential (ours)}

  \addplot[only marks, mark=*, mark size=1.5pt, rtOM] coordinates {
    (1,1.7) (2,3.4) (3,2.7) (4,3.8) (5,4.4) (8,6.2) (10,7.5) (25,17.4)};
  \addplot[only marks, mark=square*, mark size=1.4pt, rtJO] coordinates {
    (1,3.9) (2,5.9) (3,4.2) (4,4.5) (5,5.1) (8,4.8) (10,5.9) (25,12.0)};

  \addplot[rtMuted, dashed, line width=0.5pt] coordinates {(8.5,0) (8.5,6.5)};
  \node[font=\tiny, text=rtInk, anchor=west] at (axis cs:9.1,2.1) {crossover};
\end{axis}
\end{tikzpicture}
\caption{Cost of semantic alignment on identical hardware, proposals, over
S3DIS~\cite{armeni20163d} Area-5 scenes. Aligning features on the panoramas directly costs
$0.64$\,s per proposal per capture against our $0.31$\,s, so its growth rate in the number
of captures is $2.0\times$ ours. 
The two cross at $8.5$ captures and the gap widens thereafter, reaching
$1.44\times$ at $25$ captures. 
}
\label{fig:runtime_scaling}
\end{figure}

\begin{table}[t]
\centering
\tiny
\setlength{\tabcolsep}{5pt}
\renewcommand{\arraystretch}{1.0}
\resizebox{0.5\textwidth}{!}{
\begin{tabular}{lccrr}
\toprule
\multicolumn{5}{c}{\textbf{Semantic alignment, identical hardware and proposals}} \\
\midrule
\textbf{Scene} & Captures & Proposals & JOPP-3D(s) & OpenMask3D(s)\\
\midrule
office\_20  & 1  & 27  & 105  & \textbf{46} \\
hallway\_8  & 2  & 12  & 71   & \textbf{41} \\
office\_12  & 3  & 57  & 241  & \textbf{152} \\
office\_11  & 4  & 36  & 161  & \textbf{137} \\
office\_15  & 5  & 61  & 313  & \textbf{269} \\
hallway\_15 & 8  & 59  & \textbf{281}  & 368 \\
hallway\_6  & 10 & 69  & \textbf{404}  & 516 \\
hallway\_5  & 25 & 92  & \textbf{1107} & 1598 \\
\bottomrule
\end{tabular}}
\caption{Wall-clock cost of semantic alignment on one NVIDIA RTX A6000, over eight
S3DIS~\cite{armeni20163d} Area-5 scenes, using identical Mask3D~\cite{schult2023mask3d}
proposals. Lower is shown in \textbf{bold}. 
}
\label{tab:runtime_scaling}
\end{table}




\section{Additional experiments}
\setcounter{table}{0}
\renewcommand\thetable{\thesection.\arabic{table}}

\noindent\textbf{3D semantic segmentation on ScanNet200.}
Tab.~\ref{tab:scannet200_sem} evaluates JOPP-3D on ScanNet200~\cite{rozenberszki2022language}, so that our results can be read against the
large body of work that does not report on S3DIS. Every closed-vocabulary
method in the table is trained on the ScanNet200 label set, whereas JOPP-3D performs no
3D semantic training and retrieves each category by language query alone. Despite this,
JOPP-3D is second only to DITR~\cite{knaebel2025ditr} and ahead of  Concerto~\cite{zhang2025concerto}, the strongest 2D-3D
fusion models trained with full supervision. Compared against OpenMask3D, the closest
open-vocabulary method and the one sharing our proposal source, the margin is $14.0$ mIoU which is much closer than our results on S3DIS. This is because ScanNet200 is an RGB-D sequence instance-centric dataset, which is OpenMask3D's problem setting. JOPP-3D also cannot take advantage of it's wide FoV based tangential decomposition and 3D point cloud reconstruction due to the lack of panoramic captures in ScanNet200.

\begin{table}[t]
\centering
\tiny
\setlength{\tabcolsep}{5pt}
\renewcommand{\arraystretch}{1.0}
\resizebox{0.5\textwidth}{!}{
\begin{tabular}{llcc}
\toprule
\multicolumn{4}{c}{\textbf{3D Segmentation on ScanNet200~\cite{rozenberszki2022language}}} \\
\midrule
\textbf{Method} & Modality & Sup. & mIoU(\%)\\
\midrule
\multicolumn{4}{l}{\textit{Closed vocabulary}} \\
MinkUNet~\cite{choy20194d}                & 3D    & {\color{red}\textbf{full}} & 25.0 \\
PointTransformer~\cite{zhao2021point}     & 3D    & {\color{red}\textbf{full}} & 27.8 \\
PointTransformerV2~\cite{wu2022point}     & 3D    & {\color{red}\textbf{full}} & 30.2 \\
OctFormer~\cite{wang2023octformer}        & 3D    & {\color{red}\textbf{full}} & 32.6 \\
PointTransformerV3~\cite{wu2024point}     & 3D    & {\color{red}\textbf{full}} & 35.2 \\
Sonata~\cite{wu2025sonata}                & 3D    & {\color{red}\textbf{full}} & 36.8 \\
Concerto~\cite{zhang2025concerto}         & 2D+3D & {\color{red}\textbf{full}} & 39.2 \\
DITR~\cite{knaebel2025ditr}               & 2D+3D & {\color{red}\textbf{full}} & \textbf{41.2} \\
\midrule
\multicolumn{4}{l}{\textit{Open vocabulary}} \\
OpenMask3D~\cite{takmaz2023openmask3d}$^\dagger$ & 2D+3D & {\color{blue}\textbf{inst}} & 26.6 \\
\textbf{JOPP-3D (ours)}         & 2D+3D & {\color{blue}\textbf{inst}} & \underline{40.6} \\
\bottomrule
\end{tabular}}
\caption{3D semantic segmentation on the ScanNet200~\cite{rozenberszki2022language}
validation set. The best performance is shown in \textbf{bold} and the second best is
\underline{underlined}. Supervision follows the convention of the main paper:
{\color{red}\textbf{full}}: full semantic supervision, {\color{blue}\textbf{inst}}:
instance masks only. $^\dagger$: evaluated by us using the authors' released
code/checkpoints.
}
\label{tab:scannet200_sem}
\end{table}

\noindent\textbf{Sensitivity to the query wording.}
Because JOPP-3D reaches every category through a language query, 
the wording of that query is a free parameter, and it is fair to ask how much
of our result rests on it. We therefore repeat the evaluation on S3DIS~\cite{armeni20163d}
Area-5 holding the instance proposals, the tangential views, and the evaluation
protocol fixed, and varying only the text handed to the CLIP~\cite{radford2021learning}
text encoder. Results are given in Tab.~\ref{tab:prompt_robustness}.

\begin{table}[t]
\centering
\tiny
\setlength{\tabcolsep}{5pt}
\renewcommand{\arraystretch}{1.0}
\resizebox{0.5\textwidth}{!}{
\begin{tabular}{lcc}
\toprule
\multicolumn{3}{c}{\textbf{Query construction on S3DIS~\cite{armeni20163d} Area-5}} \\
\midrule
\textbf{Query construction} & 3D mIoU(\%) & mAcc(\%)\\
\midrule
\multicolumn{3}{l}{\textit{Per-scene phrasing, as used in the paper}} \\
Category name directly (As reported)                        & \textbf{80.9} & \textbf{87.0} \\
Synonym substitution               & \underline{79.8} & \underline{86.2} \\
Wrapped in ``a photo of a \{\}''   & 70.9 & 80.1 \\
\midrule
\multicolumn{3}{l}{\textit{One fixed phrase per category}} \\
Ensemble of three phrasings        & 74.4 & 82.2 \\
Extended synonym substitution      & 71.9 & 79.4 \\
Descriptive attributes             & 68.1 & 76.7 \\
\bottomrule
\end{tabular}}
\caption{Sensitivity of JOPP-3D to the wording of the language query, on
S3DIS~\cite{armeni20163d} Area-5. Proposals, views, and evaluation are
identical across all rows; only the text passed to the CLIP~\cite{radford2021learning}
text encoder changes. Synonyms substituted: sofa$\rightarrow$couch, bookcase$\rightarrow$bookshelf, whiteboard$\rightarrow$writing board; extended adds
chair$\rightarrow$seat, table$\rightarrow$desk, door$\rightarrow$doorway,
column$\rightarrow$pillar, window$\rightarrow$windowpane. Descriptive attaches an attribute
(``a wooden table'', ``a glass window''). The ensemble averages the normalized text
embeddings of three phrasings per category.}
\label{tab:prompt_robustness}
\end{table}

\begin{table*}
  \renewcommand{\arraystretch}{1.0}
  \small
  \resizebox{1.0\textwidth}{!}{\begin{tabularx}{1.2\textwidth}{llc*{13}{X}}
    \hline
    \rotatebox{65}{{\renewcommand{\arraystretch}{0.8}\begin{tabular}{l}\textbf{Setting}\end{tabular}}}
    & \rotatebox{65}{{\renewcommand{\arraystretch}{0.8}\begin{tabular}{l}\textbf{Method}\end{tabular}}}
    & \rotatebox{65}{{\renewcommand{\arraystretch}{0.8}\begin{tabular}{l}mIoU\\(\%)\end{tabular}}}
    & \rotatebox{65}{Beam} 
    & \rotatebox{65}{Board} 
    & \rotatebox{65}{Bookcase} 
    & \rotatebox{65}{Ceiling} 
    & \rotatebox{65}{Chair} 
    & \rotatebox{65}{Clutter} 
    & \rotatebox{65}{Column} 
    & \rotatebox{65}{Door} 
    & \rotatebox{65}{Floor} 
    & \rotatebox{65}{Sofa} 
    & \rotatebox{65}{Table} 
    & \rotatebox{65}{Wall} 
    & \rotatebox{65}{Window} \\
    \hline
    \hline
    \multirow{4}{*}{Closed Vocab}
    & Trans4PASS+\cite{zhang2024behind}                         & 53.6 & 0.4 & 74.4 & 65.3 & \underline{84.2} & 62.9 & 36.4 & 16.0 & 32.8 & {93.1} & 44.1 & {63.7} & 75.0 & 46.9 \\
    & SGAT4PASS~\cite{sgat4pass}                            & {56.4} & 0.7 & 74.1 & 65.9 & \underline{84.2} & 64.5 & \underline{41.2} & 19.6 & 52.7 & {93.1} & {56.9} & 58.9 & {76.4} & 44.6 \\
    & 360BEV~\cite{teng2024360bev}         & {56.5} & 0.6 & 74.6 & 65.0 & 84.0 & 62.4 & 40.3 & 18.7 & 42.2 & {93.3} & 53.9 & \underline{65.9} & 76.2 & {58.8} \\
    & PanoSAMic~\cite{chamseddine2026panosamic}$^\dagger$         & \underline{61.7} & 3.3 & 72.5 & \underline{68.7} & \textbf{85.8} & \underline{71.4} & {39.4} & \underline{39.7} & 52.0 & \textbf{96.8} & 64.2 & \textbf{71.1} & \underline{79.9} & {57.1} \\
    \hline
    \multirow{4}{*}{Open Vocab}
    & OOOPS~\cite{zheng2024open}$^\star$             & 41.1 & 0.0 & 50.7 & 44.8 & 68.8 & 51.8 & 8.1 & 0.0 & 48.2 & 71.1 & 33.6 & 47.2 & 55.4 & 54.5 \\
    & \textbf{JOPP-3D(f) (ours)}           & 52.8 & \underline{14.4} & \underline{82.4} & 55.2 & 66.9 & 46.0 & 29.2 & 19.8 & 56.0 & 80.0 & \underline{69.7} & 41.1 & 62.8 & 52.2 \\
    & SAM3~\cite{carion2025sam3segmentconcepts}$^\dagger$   & 54.2 & 10.5 & 72.9 & 61.0 & {82.1} & 64.3 & 10.8 & {11.4} & \underline{59.6} & 78.9 & 50.7 & 51.7 & 78.2 & \textbf{73.5} \\
    & \textbf{JOPP-3D (ours)}           & \textbf{70.1} & \textbf{25.8} & \textbf{83.9} & \textbf{71.7} & 81.9 & \textbf{86.6} & \textbf{45.2} & \textbf{60.2} & \textbf{76.8} & \underline{95.2} & \textbf{85.9} & {64.3} & \textbf{81.6} & \underline{66.8} \\
    \hline
  \end{tabularx}}
  \caption{Per-category results on the Stanford-2D-3D-s~\cite{2017arXiv170201105A}.}
  \label{tab:SegmentationResults}
  \vspace{-0.3cm}
\end{table*}

\begin{table*}
  \centering
  \renewcommand{\arraystretch}{1.0}
  \resizebox{1.0\textwidth}{!}{\begin{tabular}{ll*{12}{c}}
    \toprule
    \textbf{Query construction} & \rotatebox{65}{mIoU(\%)} & \rotatebox{65}{Board} & \rotatebox{65}{Bookcase} & \rotatebox{65}{Ceiling} & \rotatebox{65}{Chair} & \rotatebox{65}{Clutter} & \rotatebox{65}{Column} & \rotatebox{65}{Door} & \rotatebox{65}{Floor} & \rotatebox{65}{Sofa} & \rotatebox{65}{Table} & \rotatebox{65}{Wall} & \rotatebox{65}{Window} \\
    \midrule
    \multicolumn{14}{l}{\textit{Per-scene phrasing, as used in the paper}} \\
    Category name directly (As reported) & \textbf{80.9} & \textbf{89.0} & {77.8} & {89.7} & \textbf{89.1} & \textbf{52.4} & \textbf{69.5} & \textbf{88.2} & \textbf{94.8} & \textbf{85.6} & \textbf{82.3} & {82.3} & \textbf{75.2} \\
    Synonym substitution & 79.8 & 82.0 & 73.5 & 89.7 & 89.1 & 49.3 & 69.5 & 88.2 & 94.8 & 85.6 & 82.3 & \textbf{82.4} & 75.2 \\
    Wrapped in ``a photo of a \{\}'' & 70.9 & 89.0 & 72.4 & \textbf{90.4} & 67.7 & 39.9 & 42.8 & 81.7 & 68.2 & 85.6 & 76.2 & 82.0 & 63.5 \\
    \midrule
    \multicolumn{14}{l}{\textit{One fixed phrase per category}} \\
    Ensemble of three phrasings & 74.4 & 89.0 & \textbf{78.3} & 85.0 & 82.6 & 45.5 & 43.4 & 75.3 & 87.2 & 85.6 & 77.5 & 79.2 & 70.2 \\
    Extended synonym substitution & 71.9 & 89.0 & 76.8 & 89.7 & 78.6 & 46.4 & 27.4 & 64.9 & 94.8 & 63.6 & 68.1 & 78.3 & 63.6 \\
    Descriptive attributes & 68.1 & 47.8 & 67.2 & 86.4 & 71.9 & 39.1 & 40.5 & 63.9 & 87.2 & 85.6 & 69.8 & 82.4 & 74.2 \\
    \bottomrule
  \end{tabular}}
  \caption{Per-category 3D IoU(\%) on S3DIS~\cite{armeni20163d} Area-5 under each query construction. Proposals, tangential views, and evaluation are identical across rows.
  }
  \label{tab:prompt_classwise}
\end{table*}

The method is stable under lexical substitution: replacing category names by synonyms costs
$1.1$ mIoU, well inside the spread we would expect from any single design choice, and
extending the substitution from three categories to eight leaves the ranking unchanged. The
two constructions that do cost accuracy are informative about where the sensitivity lies.
Wrapping the queries in a generic template is not a neutral edit here, 
and the loss is grammatical rather than semantic. Attaching attributes narrows each query to a particular instance appearance, which costs recall on categories whose material varies across rooms. Conversely, averaging the text embeddings of several phrasings per category recovers most of the gap to the per-scene queries at no inference cost, which is the construction we would recommend to a practitioner who does not want to choose wording by hand. 

\begin{table}
\centering
\tiny
\setlength{\tabcolsep}{5pt}
\renewcommand{\arraystretch}{1.0}
\resizebox{0.4\textwidth}{!}{
\begin{tabular}{lcc}
\toprule
\multicolumn{3}{c}{\textbf{Views per instance on S3DIS~\cite{armeni20163d} Area-5}} \\
\midrule
\textbf{Views $K$} & 3D mIoU(\%) & mAcc(\%)\\
\midrule
$K=1$  & 63.8 & 71.0 \\
$K=2$  & 67.6 & 74.9 \\
$K=3$  & 75.0 & 81.6 \\
$K=4$  & 79.1 & 85.4 \\
$K=5$ (ours) & \textbf{80.9} & \textbf{87.0} \\
$K=6$  & \underline{80.1} & 85.8 \\
$K=7$  & 79.3 & 85.1 \\
$K=8$  & 79.8 & 85.6 \\
$K=9$  & \underline{80.1} & \underline{85.9} \\
$K=10$ & 79.0 & 84.5 \\
\bottomrule
\end{tabular}}
\caption{Effect of the number of tangential views $K$ aggregated per 3D instance, on
S3DIS~\cite{armeni20163d} Area-5, sweeping every $K$ from $1$ to $10$, peaking at $K=5$. 
}
\label{tab:k_ablation}
\end{table}

Tab.~\ref{tab:prompt_classwise} resolves the same experiment per category, which localises each effect. Substituting synonyms moves only the categories that were substituted, leaving the remaining nine within $0.1$ IoU of the reported row. The template and the descriptive variants instead perturb categories they never name: the template costs \textit{floor} and \textit{column} while leaving \textit{wall} untouched, and attaching attributes costs \textit{whiteboard} most, since a whiteboard is not well described by any single material. The ensemble is the most uniform of the fixed-phrase constructions, which is what makes it the safer default. We show some selected phrase retrievals in Figs.~\ref{fig:phrase_a} and~\ref{fig:phrase_b}.

\noindent\textbf{Number of tangential views per instance.}
Tab.~\ref{tab:k_ablation} shows the effect of $K$ on
S3DIS~\cite{armeni20163d} Area-5 with the proposals, the queries and the evaluation held
fixed, so the only variable is how many views contribute to each instance embedding. The
behaviour is non-monotonic and peaks at the value used throughout the paper. A single view is
clearly insufficient: an instance seen from one tangent is frequently truncated at the image
boundary or occluded, and the embedding inherits that partial evidence, costing performance.
Adding views up to $K=5$ recovers this, since the additional tangents contribute complementary
viewpoints of the same instance. Beyond $K=5$, the computational cost increases and we don't get much gain in the performance. 
Because the later views are, by construction, the ones in which the instance is
least visible, so their crops are dominated by surrounding context and dilute the average
rather than sharpening it. This is why we retain $K=5$ and do not treat larger $K$ as a
free improvement.

\noindent\textbf{Ceiling imposed by the proposal source.}
Coverage and instance recall describe what a proposal set contains, but not what a
segmentation built on it could achieve. We therefore measure the ceiling directly: every
proposal is given its majority ground-truth label, which is the best any labelling stage
could do, and the scene is assigned semantics from those proposals.
Tab.~\ref{tab:oracle_ceiling} reports the result over all Area-5
scenes. The distance from the ceiling to $100$ is what the proposals cost; the distance from
the reported result to the ceiling is what language alignment costs.

\begin{table}[t]
\centering
\tiny
\setlength{\tabcolsep}{5pt}
\renewcommand{\arraystretch}{1.0}
\resizebox{0.5\textwidth}{!}{
\begin{tabular}{lccccc}
\toprule
\multicolumn{6}{c}{\textbf{What the proposals to JOPP-3D permit, on S3DIS~\cite{armeni20163d} Area-5}} \\
\midrule
\textbf{Proposal source} & Coverage & Recall$_{50}$ & Ceiling & Reported & Realised\\
\midrule
Mask3D~\cite{schult2023mask3d} (S3DIS)      & 95.8\% & 88.3\% & \textbf{86.8} & \textbf{80.9} & \textbf{93.2\%} \\
SAM3D~\cite{sam3d} (ours)                   & \textbf{100\%} & \textbf{92.3\%} & 68.6 & 59.4 & 86.6\% \\
Mask3D~\cite{schult2023mask3d} (ScanNet200) & 60.6\% & 42.8\% & 50.0 & 38.3 & 76.6\% \\
\bottomrule
\end{tabular}}
\caption{Oracle ceiling per proposal source: mIoU obtained when every proposal is assigned
its majority ground-truth label. \emph{Realised} is the reported mIoU as a fraction of that
ceiling. 
Note that SAM3D leads on coverage and recall yet permits $18.2$ fewer points of mIoU,
due to its proposals spanning semantic boundaries.}
\label{tab:oracle_ceiling}
\end{table}

Three things follow. First, our language alignment is close to what the proposals allow:
JOPP-3D reaches $93.2\%$ of its ceiling, so at most $5.9$ mIoU is recoverable by improving
the alignment stage alone. Second, the JOPP-3D against JOPP-3D(f) difference of $21.5$ mIoU
is $18.2$ points of ceiling and only $3.3$ points of alignment, so it is a property of the
proposals rather than of the open-vocabulary mechanism. Third, and least expected, this is
\emph{not} caused by SAM3D missing objects: it covers every point of the scene and recovers
more instances at IoU $\ge0.5$ than Mask3D does. A proposal set can score well on both while
still capping the ceiling if its proposals straddle semantic boundaries.
This is the merging behaviour that Fig.~\ref{fig:sam3d_comparison} illustrates and that our adapted merging strategy reduces but does not remove. This remains one of the future works as pointed out in the \emph{Conclusion} section.

\newcommand{\rareRow}[4]{%
  \begin{subfigure}{0.495\linewidth}
    \centering
    \fbox{\includegraphics[width=\linewidth]{#1}}
    \caption*{\small\textbf{#3}}
  \end{subfigure}\hfill
  \begin{subfigure}{0.495\linewidth}
    \centering
    \fbox{\includegraphics[width=\linewidth]{#2}}
    \caption*{\small\textit{``#4''}}
  \end{subfigure}%
}
\newcommand{\rarePick}[3]{\rareRow{figures/suppl/rare2/picks/#1_rgb.png}%
                                  {figures/suppl/rare2/picks/#1_query.png}{#2}{#3}}
\newcommand{\rarePickB}[3]{\rareRow{figures/suppl/rare2/more_rare/picks/#1_rgb.png}%
                                   {figures/suppl/rare2/more_rare/picks/#1_query.png}{#2}{#3}}

\begin{figure}
\centering
{\small\textbf{Scene}\hfill\textbf{Query}\par}
\vspace{2pt}
\rarePick{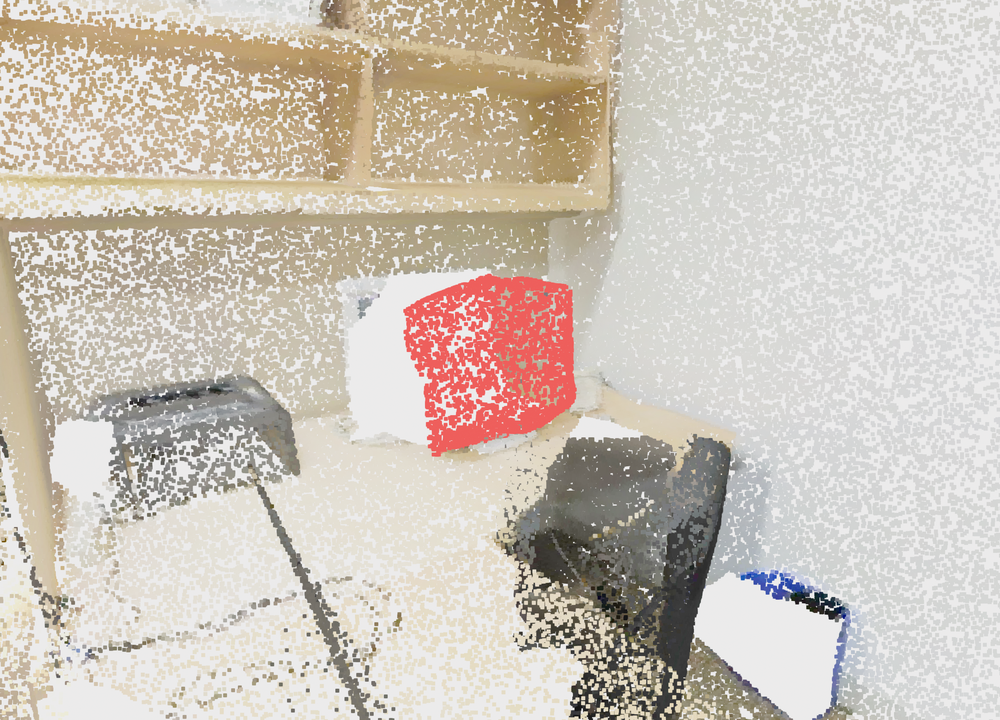}{Office}{monitor}

\vspace{3pt}
\rarePick{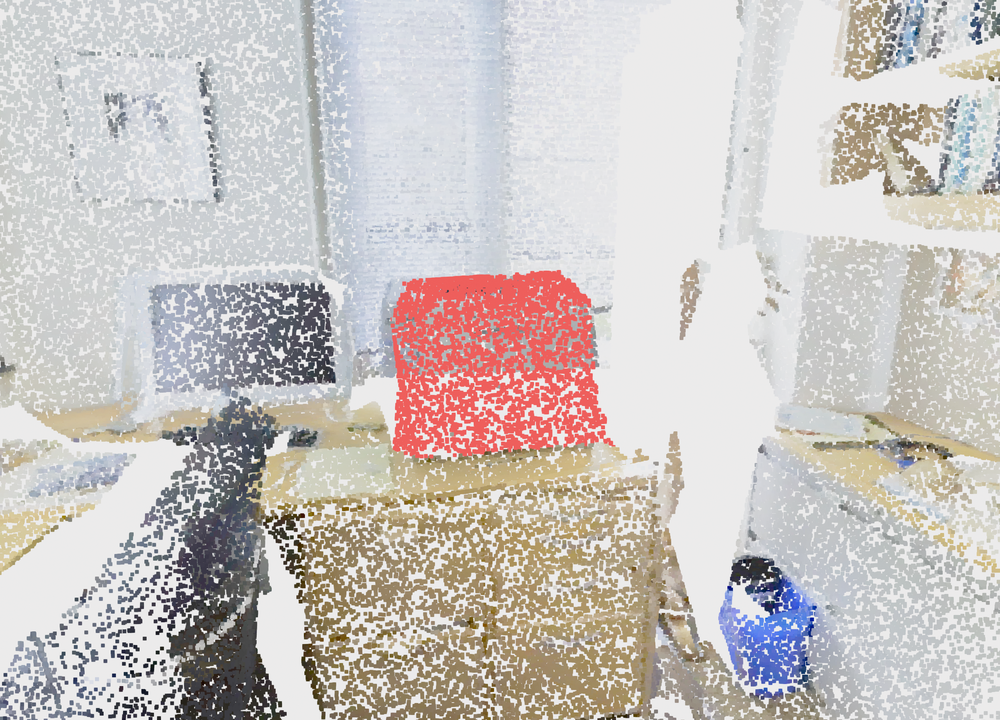}{Office}{printer}

\vspace{3pt}
\rarePick{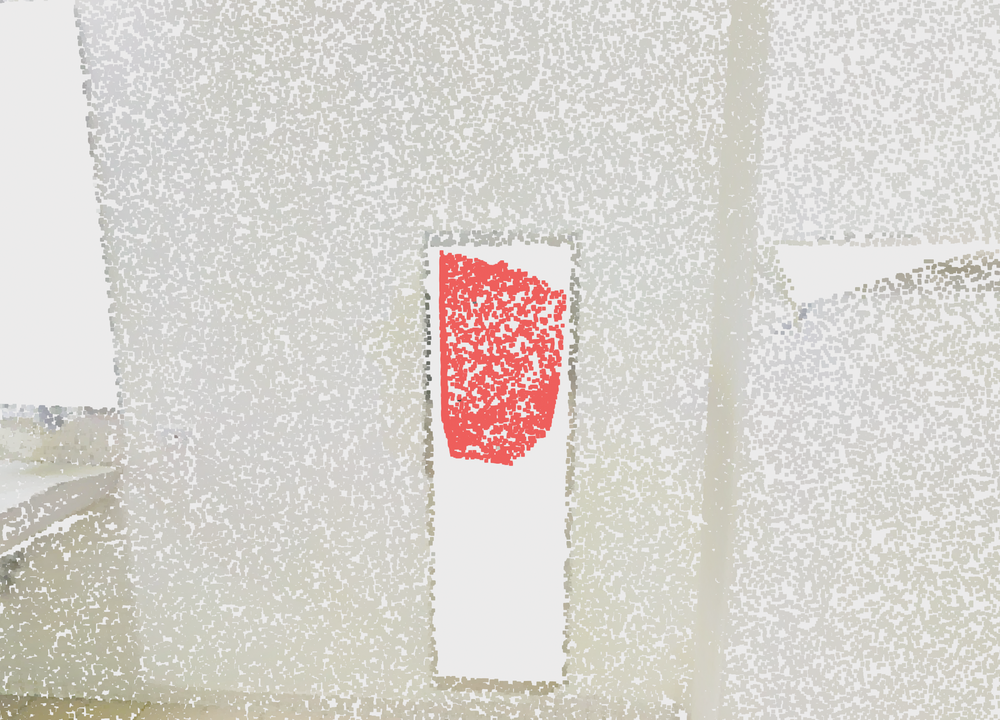}{Washroom}{hand dryer}

\vspace{3pt}
\rarePickB{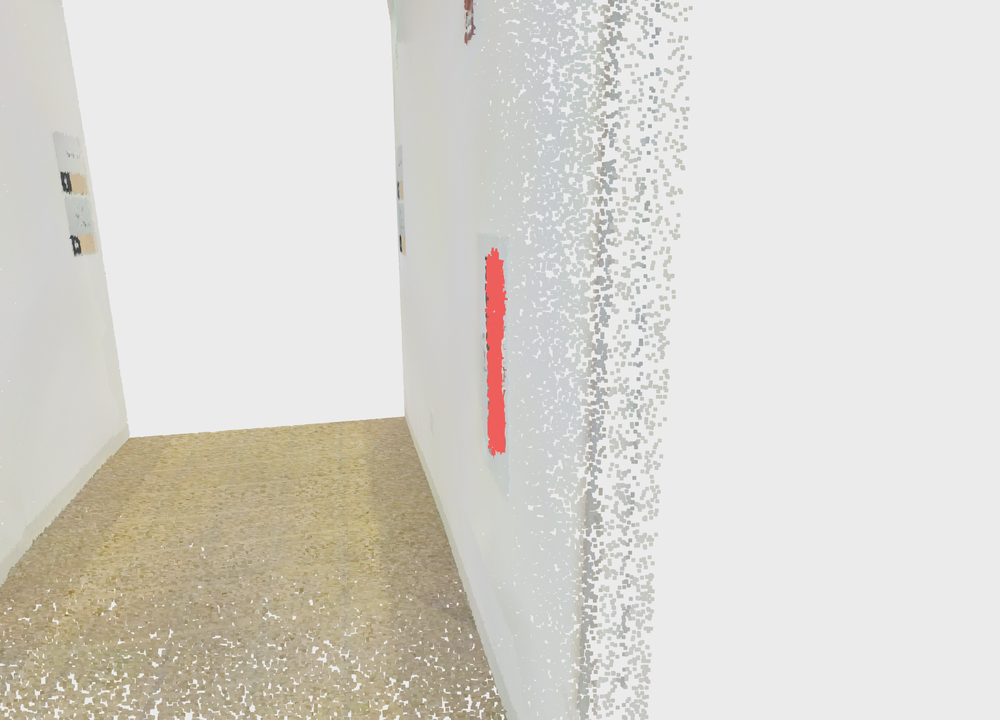}{Hallway}{fire alarm}

\caption{Additional rare-category retrievals: The retrieved
categories span office equipment, wall fixtures, personal items and greenery. None appears
in the S3DIS~\cite{armeni20163d} taxonomy, which collapses all of them into a single
\textit{clutter} label. 
}
\label{fig:rare2a}
\end{figure}

\begin{figure}
\centering
{\small\textbf{Scene}\hfill\textbf{Query}\par}
\vspace{2pt}
\rarePick{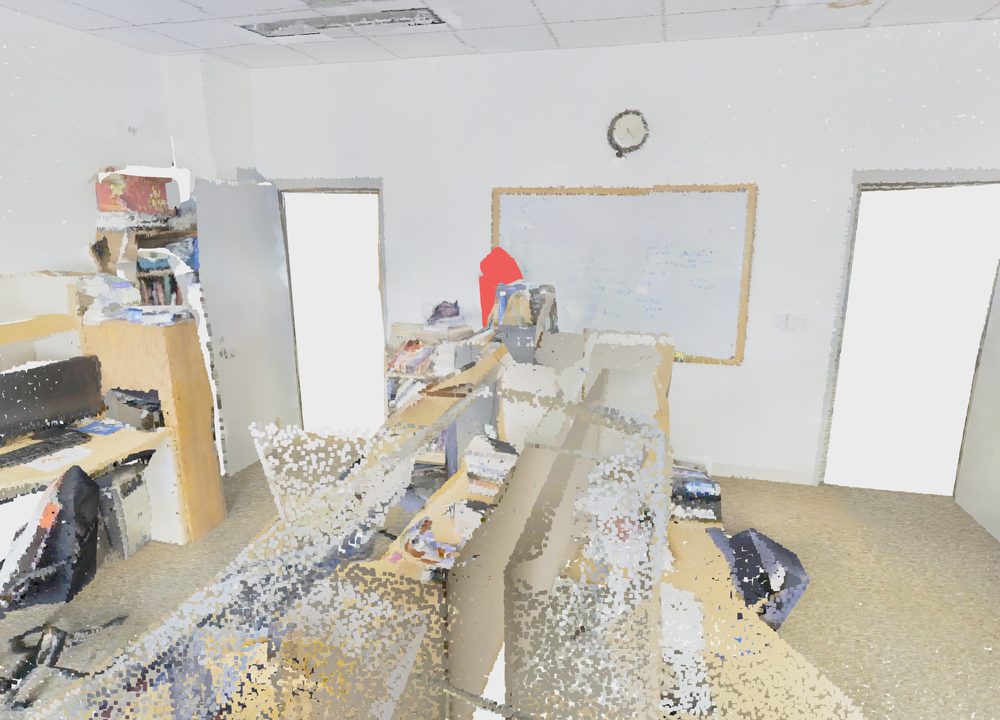}{Office}{backpack}

\vspace{3pt}
\rarePick{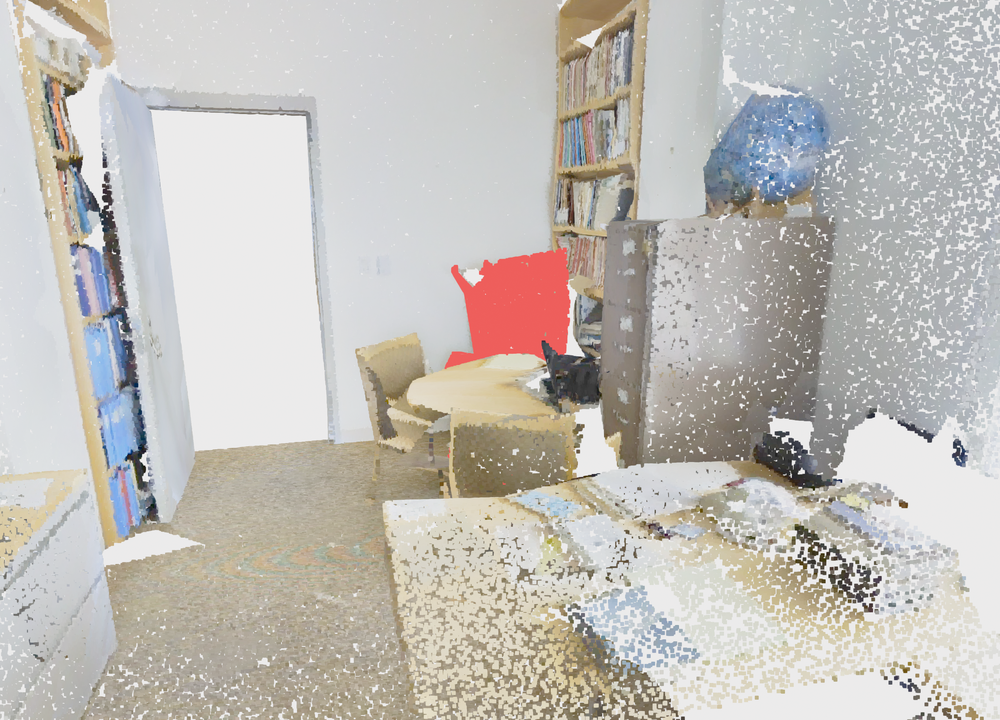}{Office}{cardboard box}

\vspace{3pt}
\rarePickB{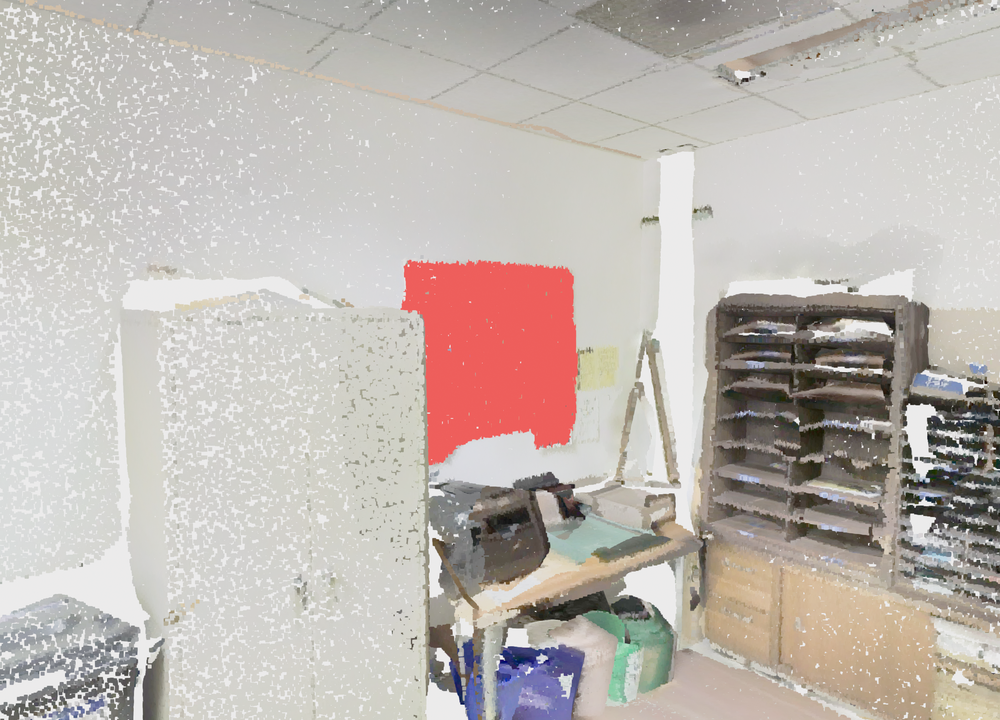}{Office}{bulletin board}

\vspace{3pt}
\rarePick{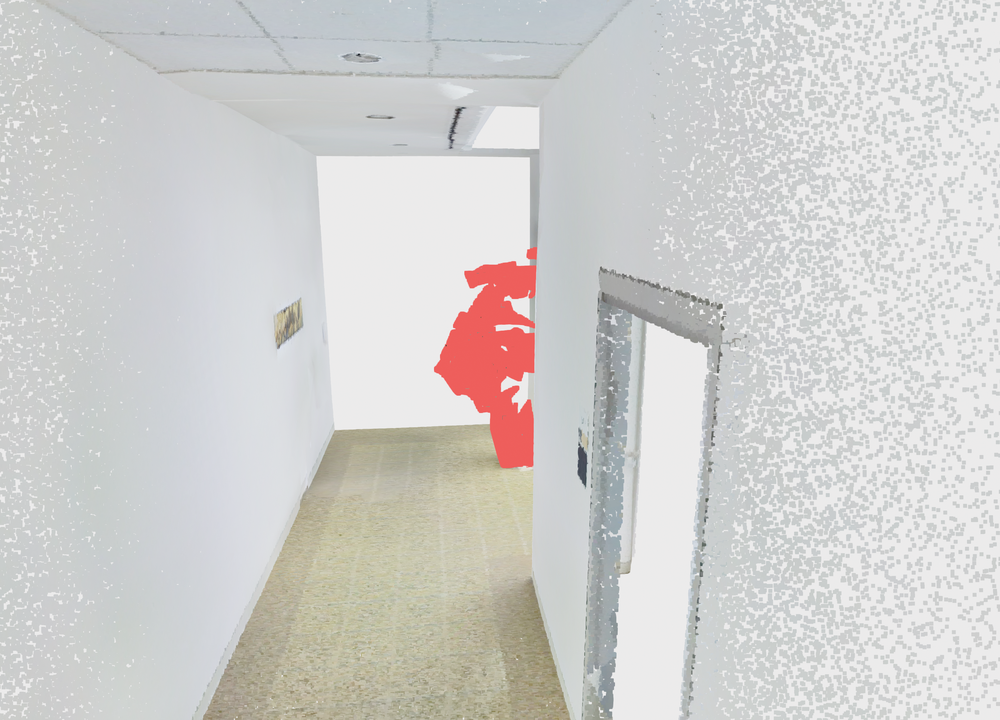}{Hallway}{plant}

\caption{Additional rare-category retrievals: continued from Fig.~\ref{fig:rare2a}.}
\label{fig:rare2b}
\end{figure}

%
\newcommand{\phraseRow}[4]{%
  \begin{subfigure}{0.495\linewidth}
    \centering
    \fbox{\includegraphics[width=\linewidth]{#1}}
    \caption*{\small\textbf{#3}}
  \end{subfigure}\hfill
  \begin{subfigure}{0.495\linewidth}
    \centering
    \fbox{\includegraphics[width=\linewidth]{#2}}
    \caption*{\small\textit{``#4''}}
  \end{subfigure}%
}
\newcommand{\phrasePick}[3]{%
  \phraseRow{figures/suppl/rare2/phrases/picks/#1_rgb.png}%
            {figures/suppl/rare2/phrases/picks/#1_query.png}{#2}{#3}}

\begin{figure}
\centering
{\small\textbf{Scene}\hfill\textbf{Query}\par}
\vspace{2pt}
\phrasePick{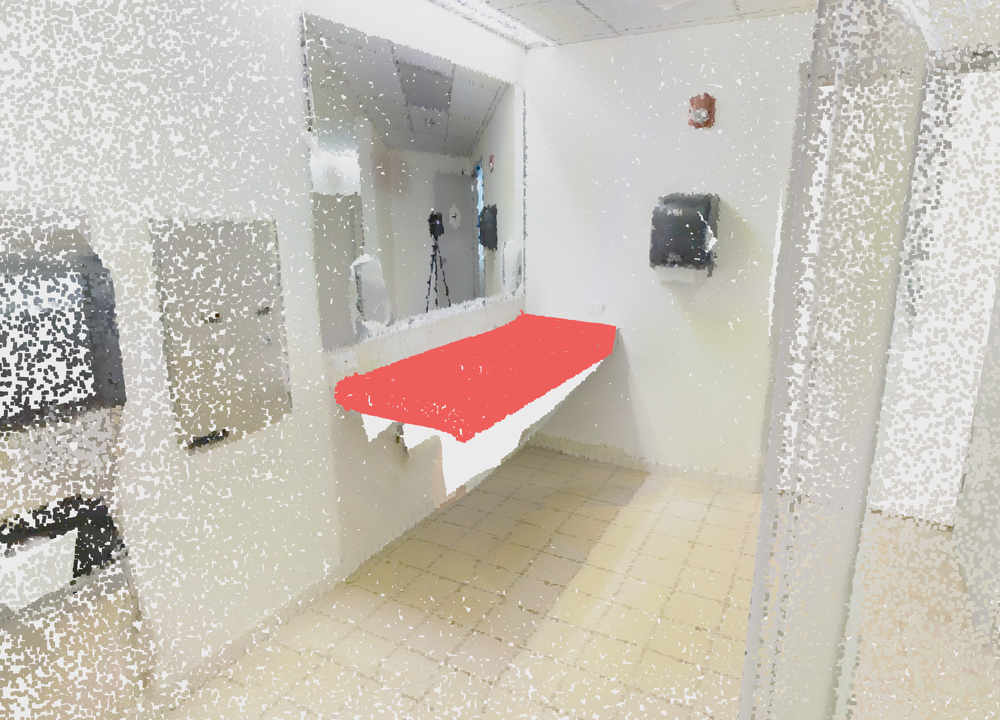}{Washroom}{a place to wash your hands}

\vspace{3pt}
\phrasePick{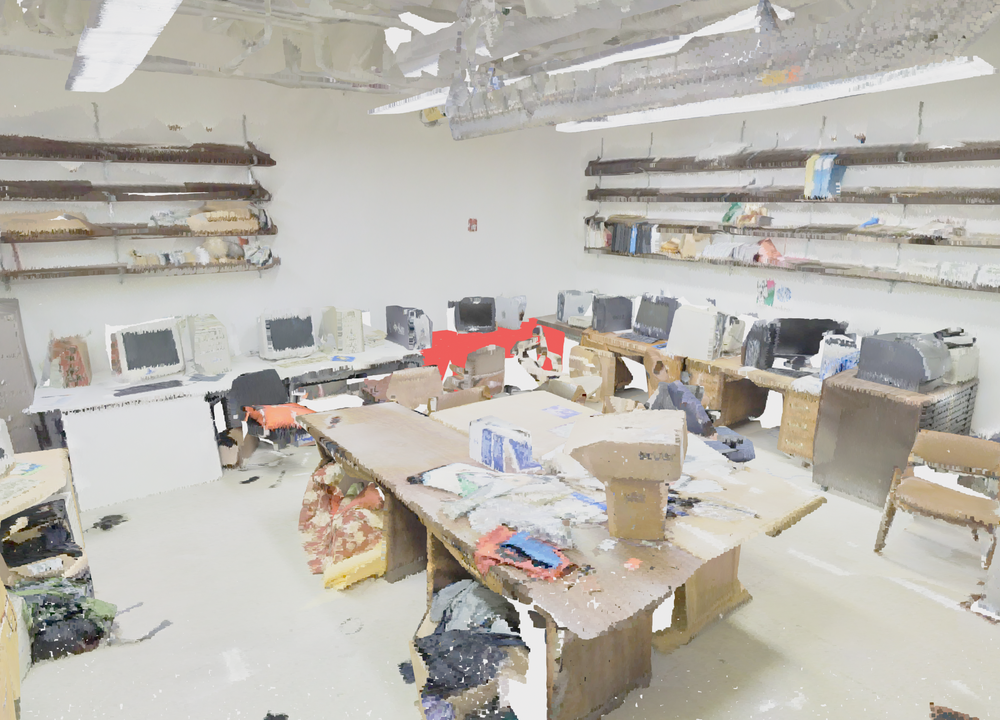}{Office}{a desk with a computer on it}

\vspace{3pt}
\phrasePick{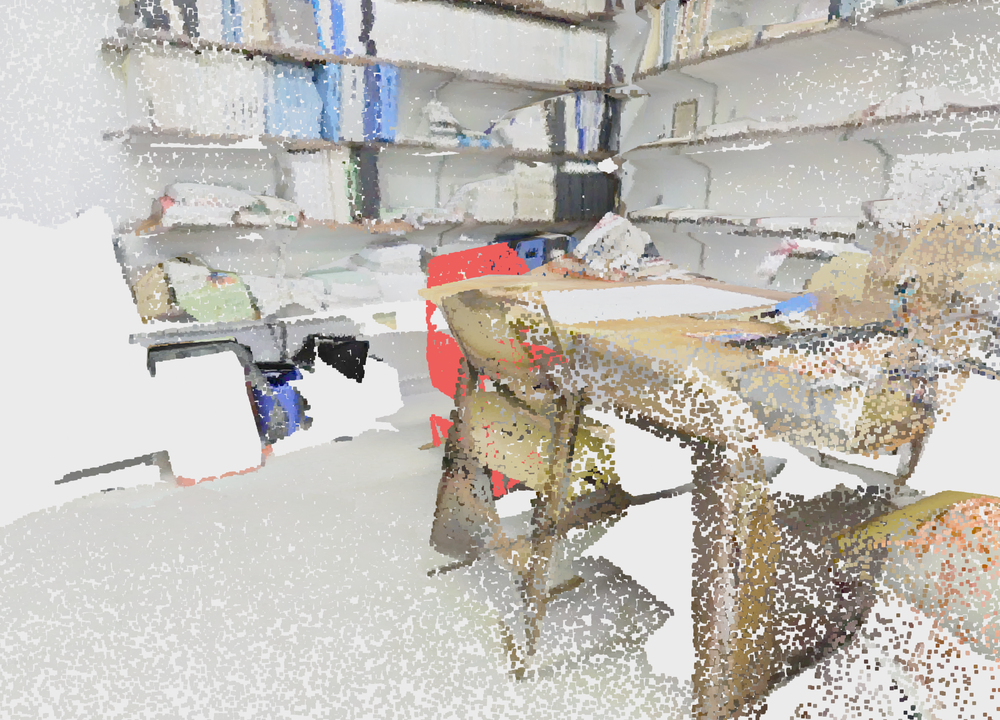}{Office}{something to sit on}

\vspace{3pt}
\phrasePick{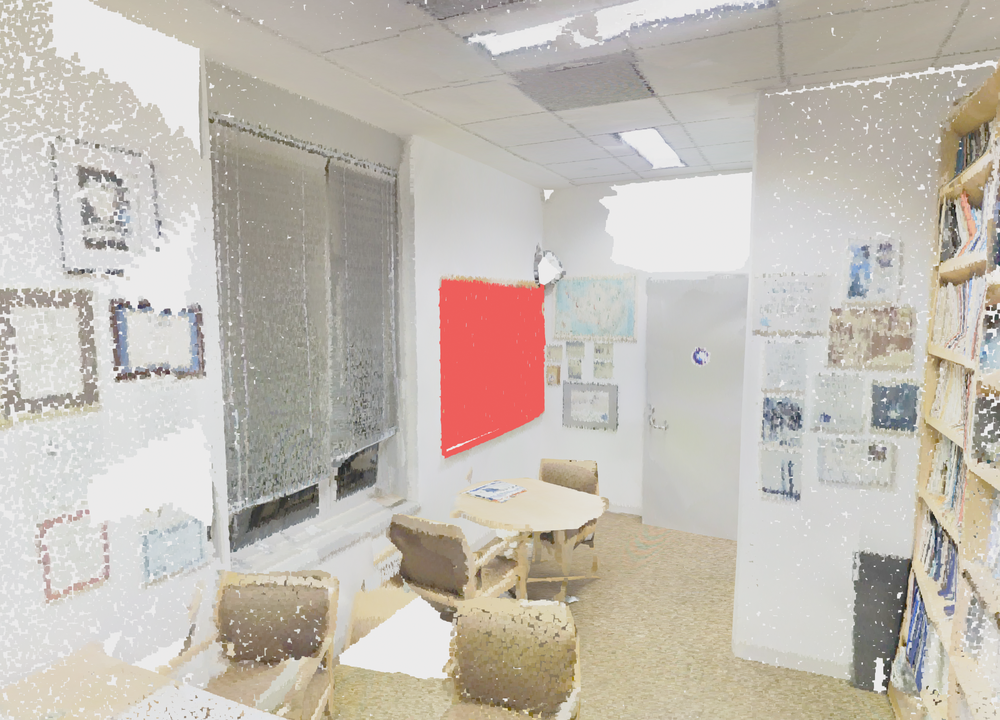}{Office}{a surface to write on}

\caption{Queries beyond category names: Each row shows
one scene: the coloured point cloud on the left and, at the same viewpoint, the instance
JOPP-3D returns for the quoted query on the right. Unlike Figs.~\ref{fig:rare2a}
and~\ref{fig:rare2b}, the targets here are ordinary annotated categories; what varies is how
they are asked for. JOPP-3D resolves functional descriptions (\textit{``something to sit
on''}) and compositional phrases (\textit{``a desk with a computer on it''}) without the
category name appearing in the query.}
\label{fig:phrase_a}
\end{figure}

\begin{figure}
\centering
{\small\textbf{Scene}\hfill\textbf{Query}\par}
\vspace{2pt}
\phrasePick{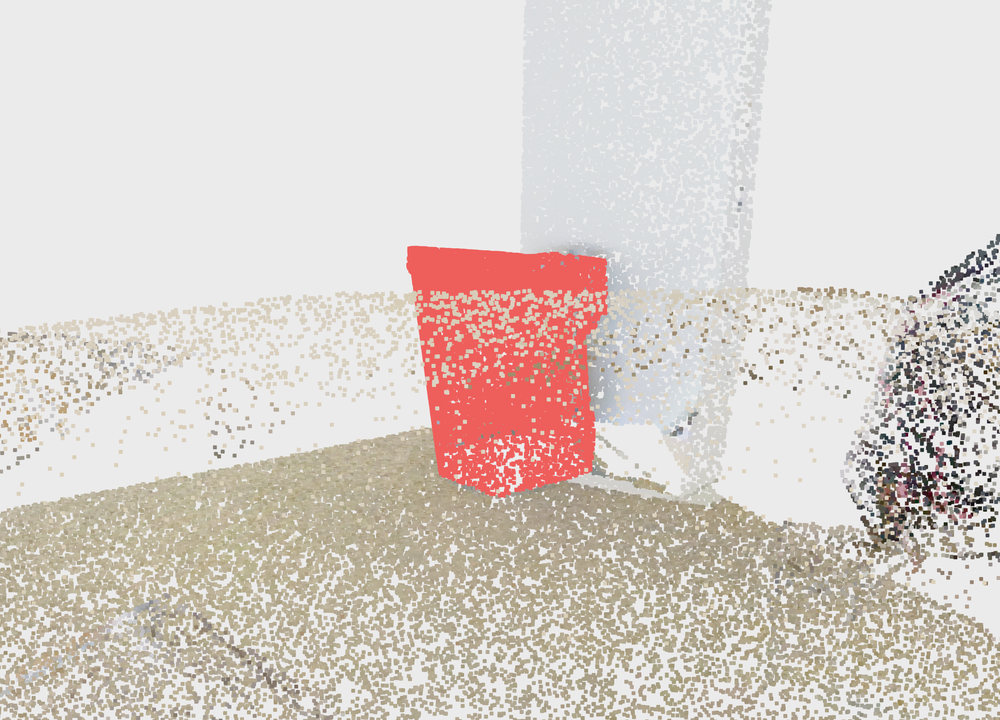}{Lobby}{something to throw trash into}

\vspace{3pt}
\phrasePick{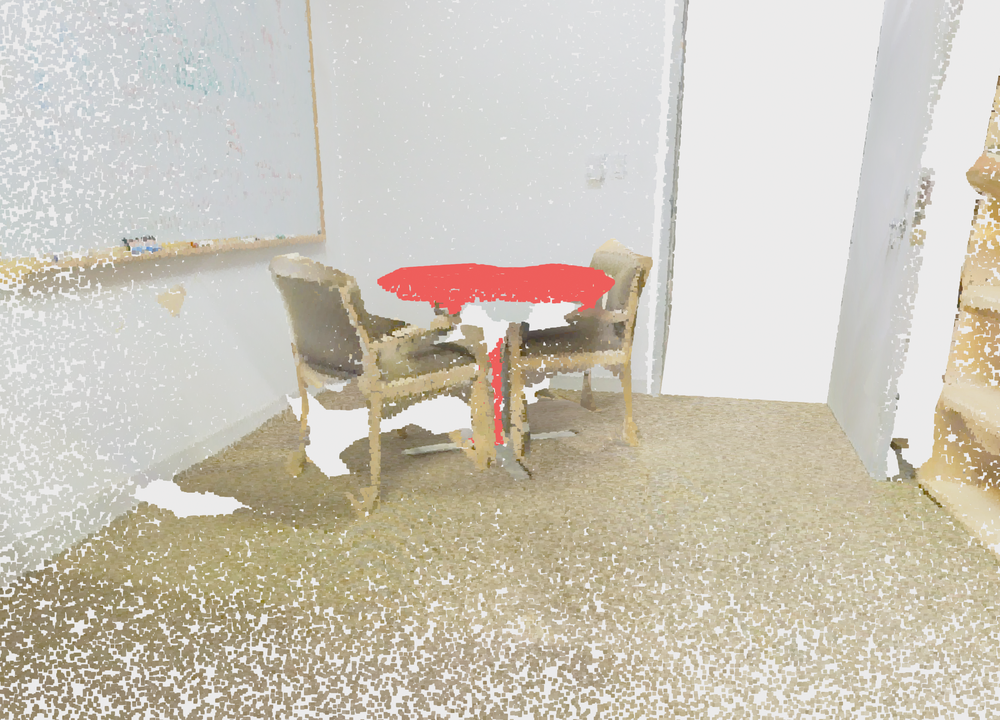}{Office}{a table with chairs around it}

\vspace{3pt}
\phrasePick{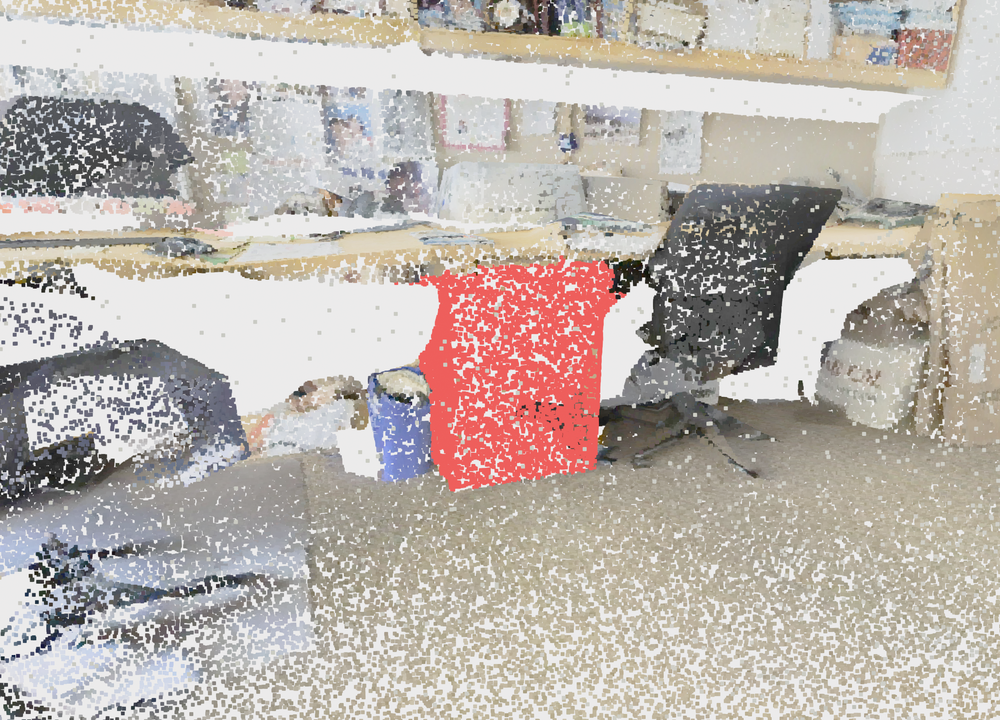}{Office}{somewhere to store files}

\vspace{3pt}
\phrasePick{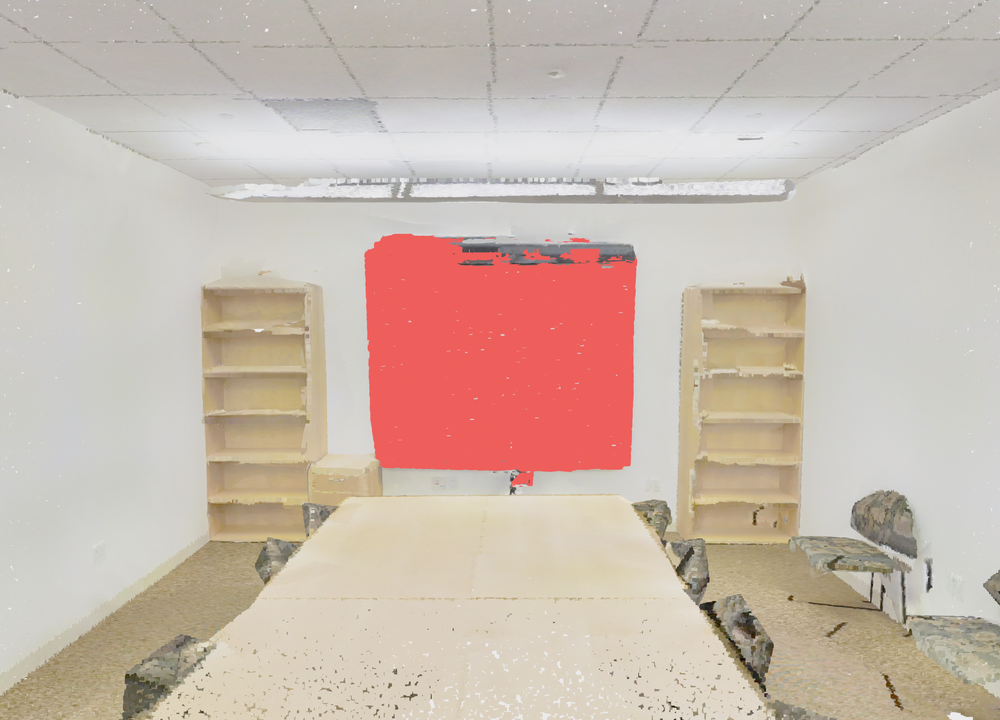}{Office}{a screen for presentations}

\caption{Queries beyond category names: continued from Fig.~\ref{fig:phrase_a}. The queries
span affordances (\textit{``somewhere to store files''}), relations between objects
(\textit{``a table with chairs around it''}) and use rather than identity
(\textit{``a screen for presentations''}). Together with the query-wording study of
Sec.~E, these indicate that the language interface is not restricted to the
benchmark's own vocabulary.}
\label{fig:phrase_b}
\end{figure}

\newpage

\end{document}